\documentclass[a4paper,fleqn]{cas-dc}
\usepackage{cite}
\usepackage{amsmath,amssymb,amsfonts}
\usepackage{algorithmic}
\usepackage{graphicx}
\usepackage{textcomp}
\usepackage{subfigure}

\def\BibTeX{{\rm B\kern-.05em{\sc i\kern-.025em b}\kern-.08em
    T\kern-.1667em\lower.7ex\hbox{E}\kern-.125emX}}
\usepackage[numbers]{natbib}

\def\tsc#1{\csdef{#1}{\textsc{\lowercase{#1}}\xspace}}
\tsc{WGM}
\tsc{QE}
\tsc{EP}
\tsc{PMS}
\tsc{BEC}
\tsc{DE}

\begin{document}
\let\WriteBookmarks\relax
\def\floatpagepagefraction{1}
\def\textpagefraction{.001}
\shorttitle{A Deep Learning Framework for  Detection of Targets  in Thermal Images to Improve Firefighting}
\shortauthors{Bhattarai et~al.}

\title [mode = title]{A Deep Learning Framework for  Detection of Targets  in Thermal Images to Improve Firefighting}                      
\tnotemark[1,2]

\tnotetext[1]{This work has been supported by NSF S\&CC EAGER grant 1637092.}

\author[1,2]{MANISH BHATTARAI}[type=editor,
                        auid=000,bioid=1,
                        orcid=0000-0002-1421-3643]
\cormark[1]
\fnmark[1]
\ead{ceodspspectrum@unm.edu}
\ead{ceodspspectrum@lanl.gov}

\credit{Conceptualization of this study, Methodology, Software}

\address[1]{Department of Electrical and Computer Engineering, The University of New Mexico, New Mexico 87106, USA}
\address[2]{Los Alamos National Laboratory,New Mexico,USA}
\author[1]{MANEL~MART\'{I}NEZ-RAM\'ON}

\cortext[cor1]{Corresponding author}

\begin{abstract}
Intelligent detection and processing capabilities can be instrumental to improving the safety, efficiency, and successful completion of rescue missions conducted by firefighters in emergency first response settings. The objective of this research is to create an automated system that is capable of real-time, intelligent object detection and recognition and facilitates the improved situational awareness of firefighters during an emergency response. We have explored state of the art machine/deep learning techniques to achieve this objective.  The goal for this work is to enhance the situational awareness of firefighters by effectively exploiting the information gathered from infrared cameras carried by firefighters. To accomplish this, we use a trained deep Convolutional Neural Network (CNN) system to classify and identify objects of interest from  thermal imagery in real time. In the midst of those critical circumstances created by structure  fire, this system is able to accurately inform the decision making process of firefighters with real-time up-to-date scene information by extracting, processing, and analyzing crucial information. With the new information produced by the framework, firefighters are able to make more informed inferences about the circumstances for their safe navigation through such hazardous and potentially catastrophic environments. 
\end{abstract}


\maketitle

\section{Introduction}
\label{sec:introduction}
The application of CNN technology abounds in the Surveillance and Defense fields  \cite{4,5,6,7} but very little research is documented in applying these principles to overcoming the navigational challenges faced by firefighters in live fire events. In fact, current firefighting modalities do not involve any automated detection mechanism and the target is identified solely by the firefighter. Detection processes can be adversely affected by environmental factors inherent in active fire scenes. High temperatures, near zero visibility caused by debris, smoke and lack of lighting, and a continuously changing environment can combine to disorient and further inhibit decision making processes, affecting even experienced firefighters. Under such hazardous conditions, lives can be lost due to rescue operation decisions based on incomplete or inaccurate understanding of the most current environmental conditions within the structure. Federal Emergency Management Agency studies \footnote{\url{https://www.usfa.fema.gov/downloads/pdf/publications/ff\_fat17.pdf}}
show a  majority of firefighter mortalities reported, resulted from inefficient decision making protocol. Heightened anxiety levels leading to misinterpretation of the scene, as well as lack of a complete understanding of the environment are cited as factors.  We propose an Artificial Neural Network-based system capable of autonomously identifying objects and humans in the scene of the event in real time to improve on-the-ground knowledge that dictates decision making protocol. The artificial intelligence (AI) based results can be used to assist in reducing these mortality statistics by minimizing anxiety induced errors. The AI-based system is also capable of accurately differentiating between human postures. This posture detection can assist firefighters in prioritization of rescue of identified victims through estimation of their health condition based on their posture. 

In this paper, we demonstrate a CNN-based autonomous system capable of generating information that can improve situational awareness for firefighters regarding the environment into which they are deployed.  The information is generated by classifying fire, objects of interest like doors, windows, and people and other thermal conditions using infrared video that is actively recorded by firefighters on scene. The CNN system can detect and classify desired targets and relay the information back to firefighters, thereby providing crucial information necessary to informing important planning decisions. This enables the firefighters and their commanders access to data collected and processed through an unbiased lens that is both comprehensive and reliable. The improved knowledge of local events and changes across the scene allows leadership to completely assess the local critical conditions and make appropriate decisions based on real time conditions. The improvement in situational awareness provided by the reliable stream of information deduced by the CNN could also assist disoriented firefighters to choose a safer path in a fire environment by autonomously identifying and alerting firefighters to the presence of objects of interest such as doors, windows, human targets, excessive smoke, etc that they may have overlooked in their confusion.

The conventional firefighting system uses different sensors such as temperature, UV, and fire detectors to determine the presence of fire, smoke, and other hazards \cite{8, 9}. Their long-established usage is evidenced by the prevalence of such sensors in all buildings, and is a requirement in building codes, to generate timely response for first responders. However, such detectors typically have a long response time in large spaces \cite{10}. Furthermore, they do not provide any spatial information regarding the presence of hazards in the given scenario. More contemporary firefighting modalities for detection of fire, smoke and other targets in a fire environment rely on color \cite{10,11,12,13}, motion \cite{14,15} and texture \cite{13,16} features of the captured image. These vision-based approaches use histogram thresholding, optical flow- based motion vector computations, and texture analysis. These more modern techniques provide enhanced performance using RGB imagery and are an improvement over the more conventional techniques described above. These algorithms do not perform as robustly on Infrared (IR) or darker imagery and also require longer computational time. IR imagery lacks sufficient complexity needed by such algorithms to perform well. Conversely, RGB images are hard to classify in active fire environments due to heavy smoke and poor lighting. IR image technology fills this gap. Furthermore, the presence of fire and smoke, by its very nature, creates a non-stationary environment and renders most existing stationary vision-based detection systems ineffectual in informing decision-making processes in real time. To address this issue, a robust real-time detection system based on CNN is proposed which is able to detect and localize the target of interest instantaneously. The research presented found in \cite{17}, describes the usage of infrared images to extract motion and statistics features in real time using a Bayesian classifier for multi-class identification. Significant research has also been done in human detection in other dark/ low visibility environments utilizing a single visible camera and fusion of the generated RGB image with an IR data set. Single IR camera based detection mechanisms have been presented in \cite{18,19,20,21}. Most of these approaches use HOG-based feature extraction and a classifier using SVM or other ML-based techniques \cite{20,22}. Paper \cite{18} presents the use of a GMM system in human detection. Template matching techniques and thresholding techniques have also been reported in \cite{20,21,23}.

Other published works that have influenced our research perform classification tasks on thermal imaging. In \cite{Cho2018}, a CNN is used in material recognition with non-firefighter grade thermal cameras. A related study is reported in \cite{Vandecasteele2017}, where transfer learning is used to detect objects in a fireground. Nevertheless, the number of data samples used in the Vandecasteele paper is low, so deep learning cannot be applied. Saliency detection and convolutional neural networks are applied to detect wildfires in \cite{Zhao2018}. 
In \cite{Kim2016} a Bayesian procedure to detect fire and smoke and discriminate them from thermal reflections in infrared is used in \cite{Yun2018}. In the recent work \cite{Ajith19}, authors introduce a methodology based on Random Markov fields to segment fire, smoke and background in a sequence of images.  

Further works that use thermal imagery and deep learning include \cite{Zhang2018}, where authors use a CNN to detect known objects in infrared surveillance cameras. In \cite{Wang2016}, a CNN is applied to the detection of vehicles in thermal imagery. Researchers in \cite{DeOliveira2018} introduce the use of CNNs to detect pedestrians in order to apply detection to unmanned aerial vehicles (UAV). Another application in UAV is presented in \cite{Rodin2018}, where authors train a CNN structure to detect objects of interest, such as bodies or body parts and other objects related to victims of avalanches.  A similar approach uses CNNs on long wave infrared imagery to detect objects. Work \cite{Valldor2014} uses deep learning to detect people with a semi-supervised approach that takes advantage of a large quantity of non-labeled images containing humans.

In spite of the large quantity of works related to the processing of infrared images in fireground or related scenarios, to our knowledge, no work has been published that attempts to construct a system that integrates online detection of targets of interest in these scenarios, including humans, objects, poses or the presence of fire,  with a large quantity of images recorded in real fire training situations by firefighters. Thus, there is a need for effective automatic target detection generated in real time in a firefighting environment as well as the associated need for a highly accurate classifier. Our research seeks to address these needs. To do so, we have adapted and enhanced an existing state of the art CNN based automatic classifier system to improve its efficacy in identifying and classifying humans and objects of interest in real time in a firefighting environment. Also, we have improved upon the creation process of a data set so that it may be used to effectively train the neural network (NN) system. We have also trained the system to detect objects and humans simultaneously. Prior technique capabilities were limited to one or the other. To further assist rescue operations, we also added a posture detection element in which the CNN further distinguishes whether a person is in a sitting, crawling, prone, or upright position. The intent of this detection series is to allow rough estimates of health condition or panic state of the victim to be approximated and used in the prioritization of rescue operations.

Panic detection work has also utilized machine learning techniques to analyze video footage, typically acquired from stationary surveillance cameras. Two papers provide excellent summaries of the techniques authored to date, focused on crowd analysis and panic detection.  \cite{panic_zhang1} covers crowd state detection methods utilizing stationary RGB video surveillance footage capable of detection at micro/macro levels to analyze local/global individual movements in the frame. The analyses usually focus on small subsets of a crowd which are then aggregated together to achieve global attributes. The analyses summarized discussed two possible approaches in their frameworks, physics-based or machine learning-based. Physics-based models use collected motion such as velocity, correlation function, fluid dynamics, energy and entropy, force model and complex systems. Machine learning-based models utilize features extracted through signal processing or computer vision tools to detect crowd state. The panic detection algorithms are applied to three different types of datasets which are generated under three situations: 1) controlled experiment 2) crowd model and simulation 3) crowd video surveillance. \cite{panic_kok}covers crowd state detection and compares a stationary crowd (the body movements of people who do not move) to a dynamic crowd (people moving from one place to another). The analysis on dynamic crowds focuses on movement patterns of the individuals in the scene to infer activity. The motion vectors detected by classical image processing techniques such as frame difference or optical flow are analyzed to deduce crowd activity. Information is then gathered by processing the frames for a dynamic crowd. The camera position is required to be fixed to obtain realistic motion features. \cite{panic_kok} found that very few studies have focused on panic detection within stationary crowds due to the challenges involved in detecting panic behaviors from small body movements such as expression or posture. However, in both crowd types, the fixed position of the camera is a requirement. These methodologies also require the analysis of a sequence of frames processed together to compute the motion vectors. In both papers, panic behavior is largely based on the rate of displacement computed for extracted features from sequences of frames and requires the camera to be stationary.

The primary focus of our work is object detection of key features like entrance/exit points (doors and windows) and persons needing rescue within an active fire environment. We also add a posture detection framework to assist in prioritization of rescue. Due to the nature of generation of our data (a handheld thermal imagery camera carried by one of the responding firefighters throughout the scene and thus non-stationary) the pose detection is limited to basic poses (sitting, standing, crawling) that are easily distinguishable in individual frame analyses and are thus not dependant upon the surrounding images in a sequence. Furthermore, these largely different poses are not reliant on high levels of detail and thus can be discerned in IR imagery. The application of the algorithms deployed in the above mentioned panic behavior detection research are difficult to deploy on our dataset because the moving camera necessary to complete our research objectives, would produce a much noisier motion vector for the targets and those movements needed to deduce panic state would be indistinguishable from those induced by the  camera movements. To minimize such errors induced by camera movements in our dataset, we process one frame at a time and infer activity rather than relying on motion vectors determined from a sequence of frames analyzed together.

Thermal imaging cameras are the most widely used cameras for obtaining footage and circumventing visibility issues within a fireground. Albuquerque Fire Department(AFD) and Santa Fe Fire Department keep at least one camera in a truck at all times. The National Fire Protection Association also supports their usage over RGB-based cameras, which are dependant on light sources for the clear depiction of imagery. In active fire scenes, where electricity may be out and no natural lighting is available, the thermal signature picked up by infrared cameras becomes crucial. Smoke and debris further reduce clarity of view for an RGB-camera but thermal cameras are able to cut through both if a heat source lies beyond the walls of smoke and dust. However, thermal imaging cameras capable of surviving the temperatures found in buildings ablaze, are costly and are not standard issue for every firefighter to carry. Instead, one per fire crew is deployed. Thus, creation of a system that can maximize the thermal imagery available, process it and return accurate, real-time information back to every firefighter on scene would dramatically enhance the combined situational awareness of the responders as a unit. Our research is restricted here to thermal imaging, but it can be extended to RGB or UV imaging as well.

\begin{figure*}
    \centering
    \includegraphics[scale=0.5]{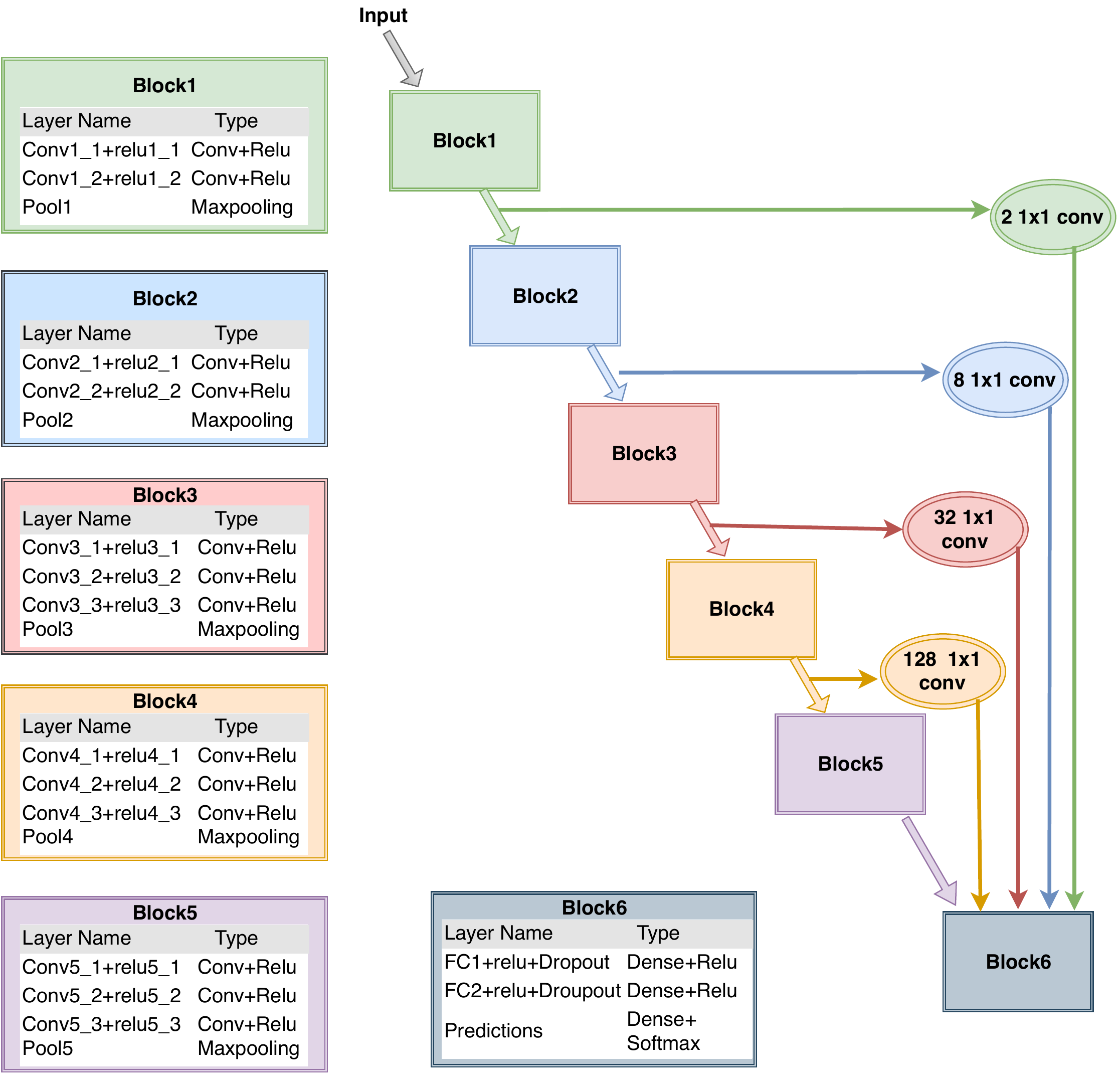} 
    \caption{Block diagram of VGG16 at different depth levels.}\label{fig:1and2}
\end{figure*}

\begin{figure*}
    \centering
    \includegraphics[scale=0.5]{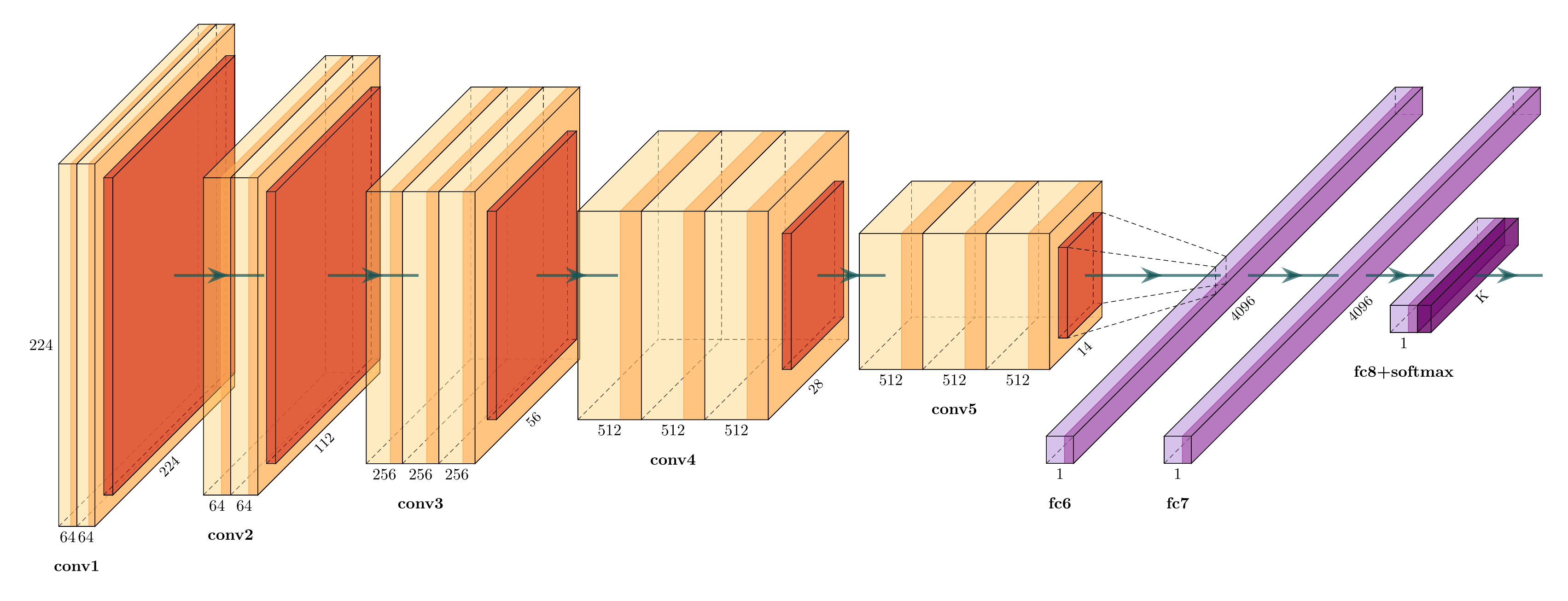} 
    \caption{Architechture of depth-5 VGG16.}\label{fig:archi}
\end{figure*}

\section{System Description}

The convolutional neural network presented is based on the structure of the VGG16 neural network presented in \cite{simonyan2014very}. Several different depths of the neural network have been tested, ranging from 1 to 5 convolutional sections as shown in Fig. \ref{fig:1and2}, all of them followed by a fully connected section of two layers. For the depth-1 configuration, the CNNs have an input of dimension $224\times 224$. The corresponding IR images are scaled to this size. They are convolved by $3\times3$ filters to produce 64 channels of dimension $224\times 224$. The resulting outputs are then passed through a set of ReLU activations. The process is repeated with an identical convolution, ReLU and  then pooled to produce 64 channels of dimension $112 \times 112$ pixels. Next, the output is passed through a $1 \times 1$ convolution to produce two channels of $112\times 112$ pixels. This is then processed through a fully connected layer with 4096 outputs, ReLU activation and dropout, a second identical fully connected layer and then a layer with 5 outputs and soft max activation that gives the classification scores across 5 different classes. 

For the depth-2 model, the first convolution layer is identical to the previous model, and after the first pool, two more convolutions are added that produce 128 channels of size $112\times 112$, reduced to 128 channels of dimension $56\times 56$. The subsequent layers have identical structure as in the 1 layer model, where the input to the first fully connected layer has 8 channels of dimension $56\times56$. 
The models with 3, 4 and 5 layers are constructed using the same methodology. The architecture for depth-5 is shown in \ref{fig:archi}.

The networks have been trained and tested in three different modalities. The first one covers object classification, which includes people, ladders, windows, doors and a combination of windows and firefighters (5 classes). The second modality comprises a pose-based classification, which includes standing, sitting and crawling (3 classes). The third modality is binary and includes the presence or absence of fire. For the modalities with 5 and 3 classes, the training was performed using a categorical cross entropy loss, and for the modality with two classes, we used binary cross entropy loss combined with stochastic gradient descent. The learning rate for all trainings was $10^{-4}$ with a decay of $0.009$. Early stopping with cross validation was applied to avoid overfitting. 

In order to avoid overfitting between training and test datasets, images taken from different recordings have been used in both processes. The results shown below contain extensive tests using a database of recordings taken by the researchers of this paper. It is described below.


\section{Training and Test Datasets}

The imagery data set used in this project was recorded at the Santa Fe Firefighting Facility, located in Santa Fe, New Mexico.  Extensive video footage was acquired using an IR MSA 5200HD2TIC Camera. This camera is a multipurpose firefighting tool designed to aid search and rescue efforts in structural firefighting environments. It uses an uncooled microbolometer vanadium oxide(Vox) detector which comprises of 320x240 FPA with the pitch of 38$\mu m$ and spatial resolution of 7.5 to 13.5$\mu m$. This resolution is sufficient to capture necessary features for target detection. It records the image with a 320x240 focal plane array sensor and has the ability to record imagery in two different modes, i.e. low and high sensitivity. This device also features high score imagery, generating 76,000 pixels of image detail in both low and high sensitivity modes. Dense spectral resolution is (7.5 to 13.5$\mu m$). The output video is in NTSC format with a frame rate of 30 frames per second. The scene temperature has a maximum operating range of 560 degrees Celsius or 1040 degree Fahrenheit.  

Over 6 hours of recorded video in both open and closed environments was acquired. The recording sessions produced more than  150 infrared video files, each one lasting approximately 2 to 3 minutes. All videos contained some combination of sequences involving the desired targets to be detected.  In some scenarios, single objects of interest are present in a scene while in others multiple objects of interest are present simultaneously. This variation requires the CNN image classification system to be capable of multiple simultaneous object detections in the same frame. Other objects or postures of interest outside of what we chose can also be detected if those objects or poses occur in sufficient frequency to allow for training the data.

The objects of interest for this research are humans, doors, windows, ladders, and fire. The objective of the above described structure is to detect all objects of interest present in the scene at the same time.  
\begin{figure}
    \centering  
    \includegraphics[scale=0.5]{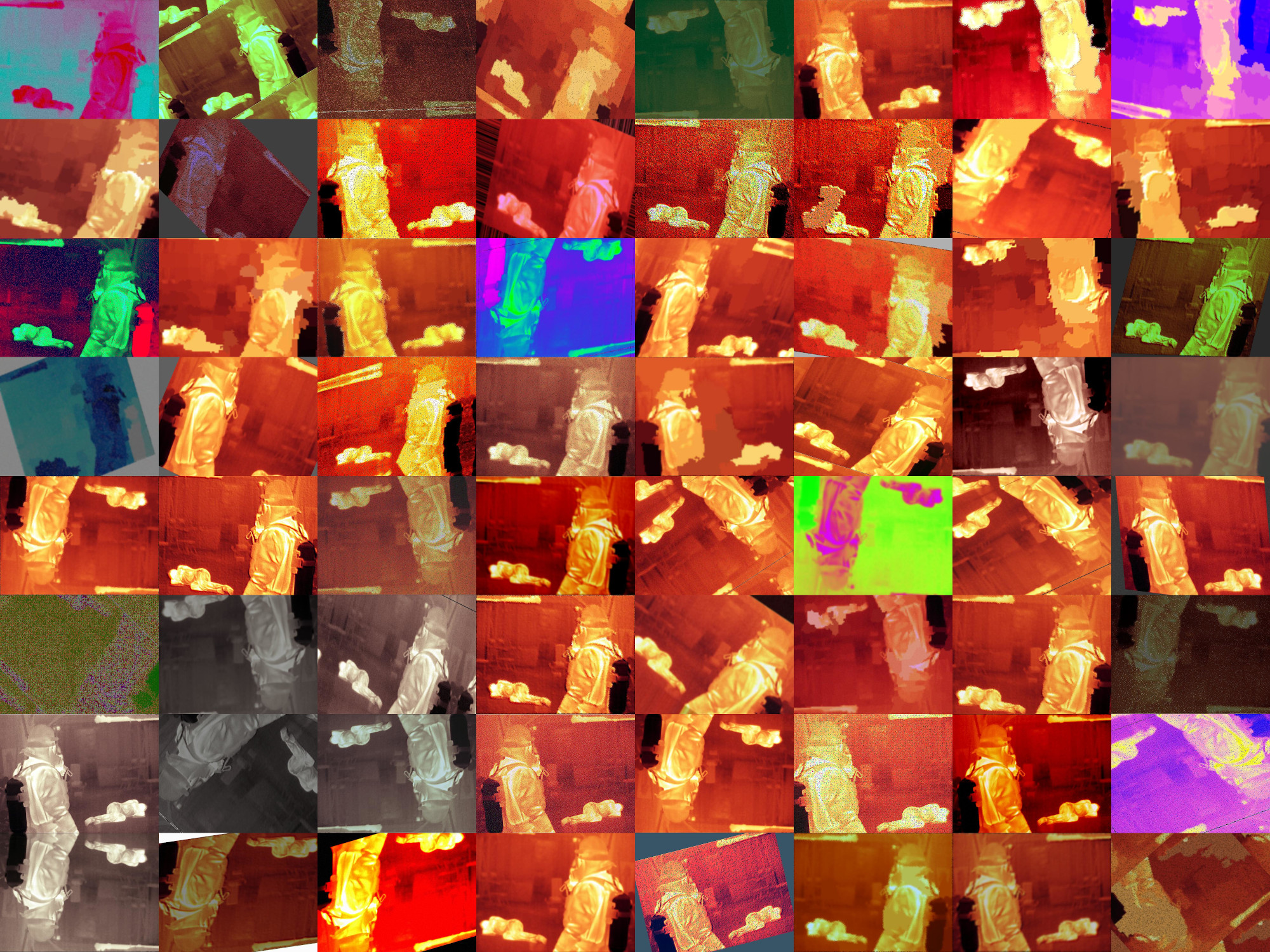} 
    \caption{Demonstration of Image augmentation for training set.}\label{fig:augment}
\end{figure}
Since data sets of sufficient detail are needed to accurately train the neural network, the videos were used to extract a large quantity of images for training and test purposes. The training and test sequences were extracted from different videos in order to avoid overfitting. The video extracted from the camera was produced in grayscale 8 bit format. In order to generate the training data set, the images were pre-processed with data augmentation techniques such as skewing, translation, zooming, cropping and rotation as shown in fig \ref{fig:augment} (False color for better visualization). 

Tables \ref{tab:data_objects}, \ref{tab:data_fire} and \ref{tab:data_poses} show the total number of images acquired for training and test before the augmentation procedure. Objects of interest contained in images used in the training data set were hand labeled to assign them to a class. The labeled classes were then grouped into 3 sets (objects, fire, human poses). To compensate for the asymmetry of data within the different classes, data augmentation was performed on classes that had lower representation within the original dataset. For example, 7950 images from the original dataset were labeled and added to the "No fire" class within the set "Fire". The "fire class" was augmented to add 347 images to it to bring the total number of "fire" labeled images available for training up to 7950. 

The primary training was performed using an Alienware Aurora R6 Desktop computer configured with 32GB RAM memory, and a Dual GTX 1080 with 16GB GPU memory.  The cross validation and hyper parameter tuning portion of the research was performed on the high performance computer, Xena housed at the UNM Center for Advanced Research Computing(CARC). The machine has 24 single GPU nodes and 4 dual GPU nodes. These dual GPU nodes have 2 NVIDIA Tesla K40m GPUs with GPU memory of 11GB each and 64GB RAM memory per node. The computations for cross validation and hyper parameter tuning utilized the 4 dual GPU nodes.

\begin{table}[]
\begin{subtable}
\centering
\begin{center}
\begin{tabular}{ |c|c|} 
\hline
Object of Interest & number  \\
\hline
door & 322 \\
firefighter and window & 4663 \\
firefighter & 15484 \\
ladder & 1589 \\
window & 1620 \\
\hline
\end{tabular}
\caption{Data quantification for objects classification task(object set)}\label{tab:data_objects}
\end{center}
\end{subtable}

\begin{subtable}
\centering
\begin{center}
\begin{tabular}{ |c|c| } 
\hline
Object of Interest & number \\
\hline
fire & 7603  \\
No fire &7950 \\
\hline
\end{tabular}
\end{center}
\caption{Data quantification for fire classification task(fire set)}\label{tab:data_fire}
\end{subtable}

\begin{subtable}
\centering

\begin{center}
\begin{tabular}{ |c|c| } 
\hline
Object of Interest & number \\
\hline
Crawling & 8678 \\
sitting & 1803 \\
standing & 9928 \\
\hline
\end{tabular}
\end{center}
\caption{Data quantification for poses classification task(poses set)}\label{tab:data_poses}
\end{subtable}
\end{table}

The test data consists of 1/10 of the data set. The rest of the data has been used for training and validation purposes. A validation has been performed with a 9-stratified fold procedure with the remaining 9/10ths of data.

\begin{figure*}
\begin{center}
\includegraphics[scale=0.53]{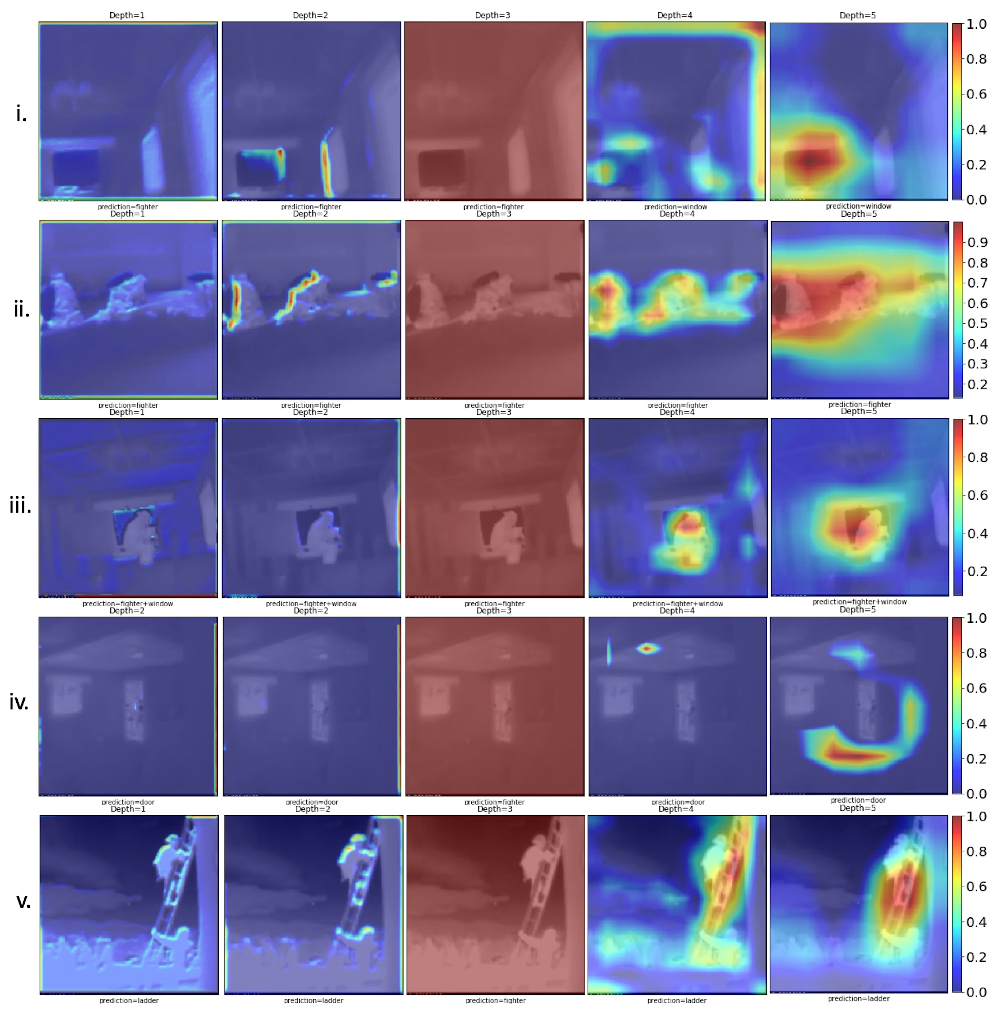}
\caption{Visualization of the features obtained at the last Convolutional layer for Object Classification for different depths for respective classes \textbf{i.} Window, \textbf{ii.} Fighter, \textbf{iii.} Fighter+Window,  \textbf{iv.} Door and  \textbf{v.} Ladder }\label{fig:features_objects}
\end{center}
\end{figure*}

\begin{figure*}
\begin{center}
\includegraphics[scale=0.53]{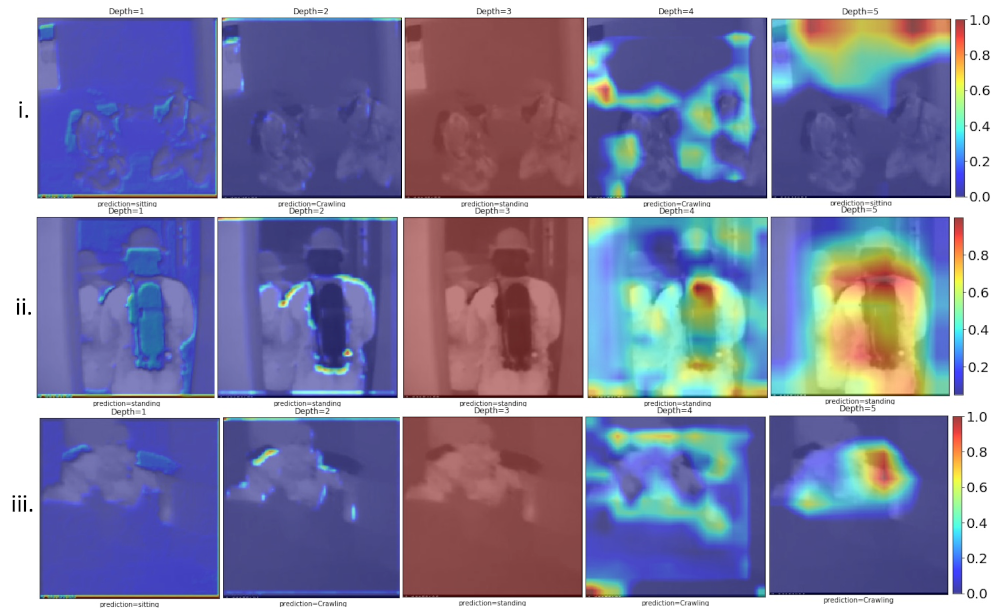}
\caption{Visualization of the features obtained at the last Convolutional layer for pose Classification for different depths for respective classes \textbf{i.} Sitting, \textbf{ii.} Standing and \textbf{iii.} Crawling} \label{fig:features_poses}
\end{center}
\end{figure*}

\begin{figure*}
\begin{center}
\includegraphics[scale=0.53]{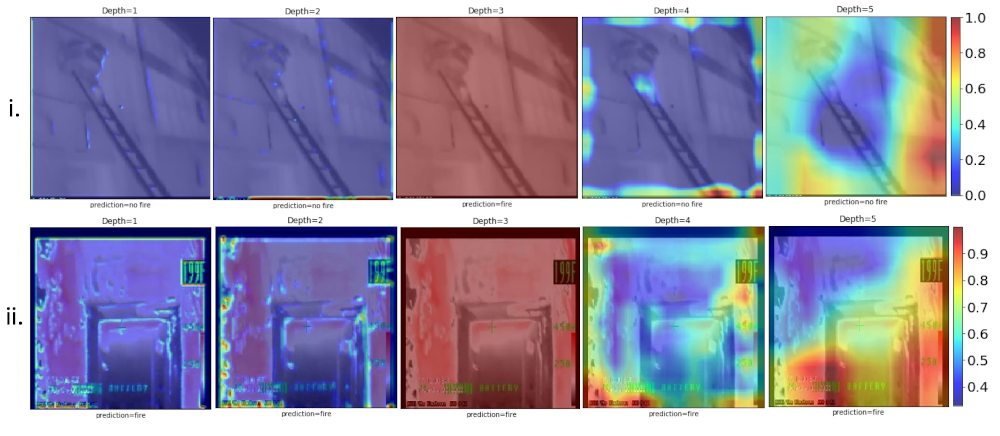}
\caption{Visualization of the features obtained at the last Convolutional layer for fire Classification for different depths for respective classes \textbf{i.} fire and  \textbf{ii.} No Fire} \label{fig:features_fire}
\end{center}
\end{figure*}

\begin{figure}[]
\begin{center}
\includegraphics[scale=0.2]{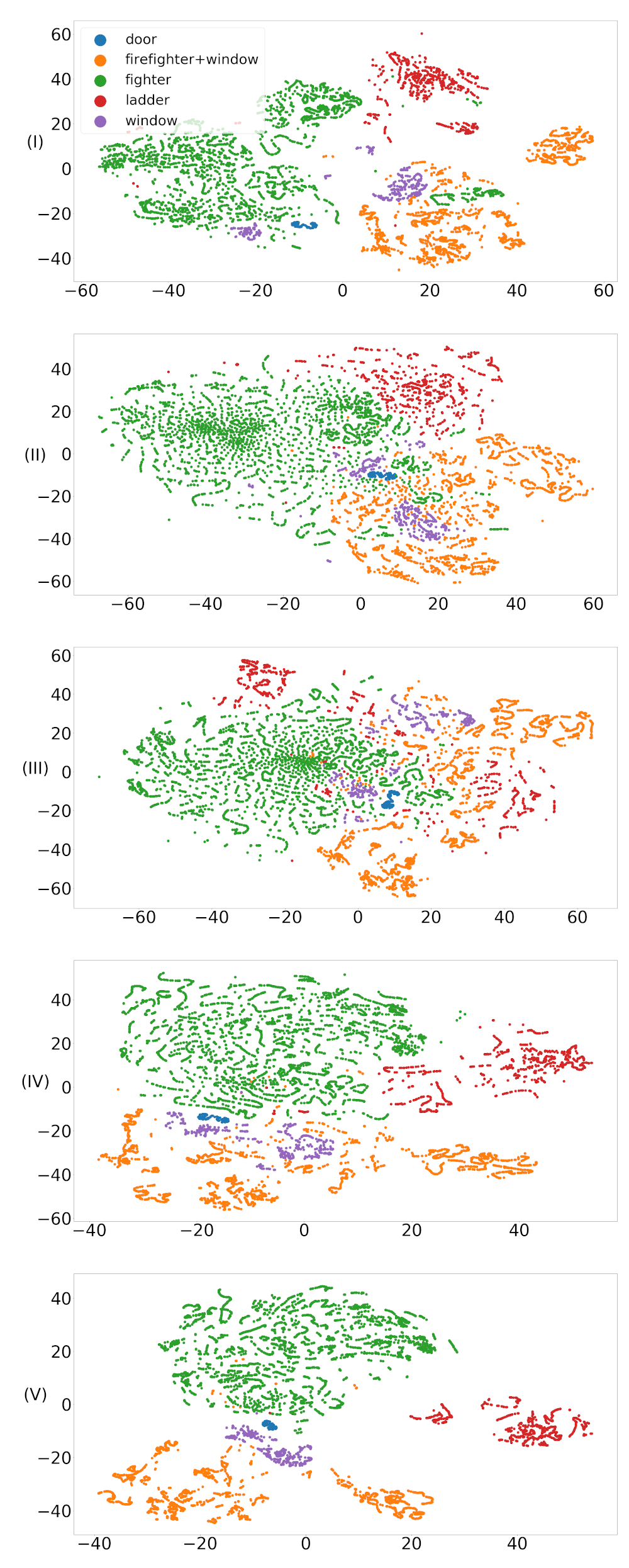}
\caption{t-SNE of the CNN when trained to detect objects. The number of convolutional layers was respectively I) depth-1, II) depth-2, III) depth-3, IV) depth-4 and V) depth-5.}
\label{fig:TSNE}
\end{center}
\end{figure}

\section{Results}

Our network is trained to detect objects including ladders, doors, windows, people, and fire. The network is also trained to detect and classify different body positions of the detected people, civilian and firefighter alike. We classify three different poses corresponding to standing, sitting and crawling, which cover the variation found in the videos. Other important positions can be detected if they are sufficiently available in the image dataset used for training.  The following results present the visualization of the features extracted by the convolutional section of the network, the F1 scores and achieved accuracy of the network, confusion matrices and ROC curves to estimate the false alarm versus the detection probability of the network. For this experiment, we change the detection probability by sweeping the detection threshold of the network.

\subsection{Visualization of the extracted features}
All the configurations were tested against the available data. We have used grad-CAM(gradient weighted Class Activation Maps) \cite{selvaraju2017grad} for visualizing attention over input. Grad-CAM uses the  Convolutional layer before the dense layer in order to utilize spatial information for displaying the saliency maps corresponding to each predicted class. 
The activations correspond to the class with the highest detection score. We can see the classification scores for different classes corresponding to different images at different depths in table \ref{tab:PredictionObjects}, \ref{tab:PredictionPoses} and \ref{tab:PredictionFire}. Based on the classification score of the table, we visualize the activations corresponding to classes with highest score.  Figure \ref{fig:features_objects},\ref{fig:features_poses} and \ref{fig:features_fire} show grad-CAM for different depths for each classification task. Each of these figure comprises of subfigures corresponding to each classes depicting the grad-CAM for different depths from left to right. For most of these maps for depth-1 and depth-2 show that activations come from either edges or small regions from  the objects of interest. With depths equal to 3, the network is unable to  produce feature extraction, showing that this configurations is not useful for classification. Nevertheless, at higher levels of abstraction (depths equal to 4 and 5),features are extracted that produce exemplary results.

Figure \ref{fig:TSNE} shows the visualization of the output of the neural networks using the t-distributed stochastic neighbor embedding technique presented in \cite{maaten2008visualizing}. This technique allows the user to obtain a low dimensional representation of the data to better understand the data's distribution and separability. \ref{fig:TSNE} (I) shows the distribution of the 5-dimensional output of the network when the depth is 1. In this configuration, the separability of the data is intuitively fair. \ref{fig:TSNE} (II) and (III) show the distribution for depth 2 and 3. These representations show a high level of overlapping of the different classes, which is consistent with the poor extraction of features, as shown in figure\ref{fig:features_objects},\ref{fig:features_poses} and \ref{fig:features_fire}. The representation for  depth of 4 and 5 as shown in \ref{fig:TSNE} (IV) and (V) is highly improved, as the overlapping is dramatically decreased compared to the 2 and 3 depth networks. 

The green dots correspond to images of firefighters. Orange dots are images containing firefighters and windows. These two classes comprise the majority of the data set but they show a low overlap between them indicating high accuracy in the classifier's ability to distinguish between firefighters in the presence or absence of a window. The blue dots correspond to images with doors. Door shapes vary due to the angle of the camera. However, the classifier is able to extract all door features regardless of these differences in perspective-induced shape, and all features appear clustered in a small area. They are highly overlapped with the images containing firefighters at lower depths. The overlap decreases with higher level of abstraction (4 and 5 depths). The violet spots represent windows, which appear to be overlapping with the images of firefighters and windows. The accuracy and classification rates are in high agreement with this visualization.

\begin{table*}[]
\centering
\caption{Prediction score for object classification for different depths (green and yellow cells are the top 2 prediction scores)}
\label{tab:PredictionObjects}
\begin{tabular}{|l|l|r|r|r|r|r|}
\hline
\cellcolor[HTML]{C0C0C0} Figure & \cellcolor[HTML]{C0C0C0} Depth & \cellcolor[HTML]{C0C0C0} Door & \cellcolor[HTML]{C0C0C0} F/W & \cellcolor[HTML]{C0C0C0} Ladder & \cellcolor[HTML]{C0C0C0} Window & \cellcolor[HTML]{C0C0C0} Fighter \\ \hline
\ref{fig:features_objects} i.	&1&	0.003&	0.053&	0&	\cellcolor{yellow!50}0.408 	& \cellcolor{green!25}0.536  \\ \hline
	&2&	0&	\cellcolor {yellow!50}0.43&	0&	0.01	& \cellcolor{green!25}0.56  \\ \hline
	&3&	0&	0&	0&	0&	 \cellcolor{green!25}1  \\ \hline
	&4&	0.002&	\cellcolor {yellow!50}0.037&	0&	 \cellcolor{green!25}0.961&	0  \\ \hline
	&5&	0.001&	\cellcolor {yellow!50}0.004&	0&	 \cellcolor{green!25}0.996&	0  \\ \hline \hline
\ref{fig:features_objects} ii.	&1& \cellcolor {yellow!50}0.006&	0&	0.001&	0.001&	 \cellcolor{green!25}0.992  \\ \hline
	&2&	0&	0&	0&	0&	 \cellcolor{green!25}1  \\ \hline
	&3&	0&	0&	0&	0&	 \cellcolor{green!25}1  \\ \hline
	&4&	0&	0&	0&	0&	 \cellcolor{green!25}1  \\ \hline
	&5&	0&	0&	\cellcolor {yellow!50}0.001&	0&	 \cellcolor{green!25}0.9999  \\ \hline \hline
	
\ref{fig:features_objects} iii.	&1&	0&	 \cellcolor{green!25}0.997&	0&	\cellcolor {yellow!50}0.003&	0  \\ \hline
	&2&	0&	 \cellcolor{green!25}1&	0&	0&	0  \\ \hline
	&3&	0&	0&	0&	0&	 \cellcolor{green!25}1  \\ \hline
	&4&	0&	 \cellcolor{green!25}0.974&	0&	\cellcolor {yellow!50}0.025&	0  \\ \hline
	&5&	0&	 \cellcolor{green!25}0.999&	0&	\cellcolor {yellow!50}0.001&	0  \\ \hline \hline

\ref{fig:features_objects} iv.	&1&	 \cellcolor{green!25}0.988&	0.001&	0&	0.001&	\cellcolor {yellow!50}0.01  \\ \hline
	&2&	 \cellcolor{green!25}0.996&	0&	0&	0.001&	\cellcolor {yellow!50}0.003  \\ \hline
	&3&	0&	0&	0&	0&	 \cellcolor{green!25}1  \\ \hline
	&4&	 \cellcolor{green!25}0.95&	\cellcolor {yellow!50}0.029&	0&	0.02&	0.001  \\ \hline
	&5&	 \cellcolor{green!25}0.805&	0.01&	0&	\cellcolor {yellow!50}0.181&	0.004  \\ \hline \hline

\ref{fig:features_objects} v.	&1&	0.007&	0.001&	 \cellcolor{green!25}0.976&	0.001&	\cellcolor {yellow!50}0.015  \\ \hline
	&2&	0&	0&	 \cellcolor{green!25}0.999&	0&	\cellcolor {yellow!50}0.001  \\ \hline
	&3&	0&	0&	0&	0&	 \cellcolor{green!25}1  \\ \hline
	&4&	0&	0&	 \cellcolor{green!25}0.998&	0&	\cellcolor {yellow!50}0.002  \\ \hline
	&5&	0&	0&	\cellcolor{green!25}0.9&	0&	 \cellcolor{yellow!50}0.1  \\ \hline 

\end{tabular}
\end{table*}

\begin{table}[]
\centering
\caption{Prediction score for Pose classification for different depths}
\label{tab:PredictionPoses}

\begin{tabular}{|l|l|r|r|r|}
\hline
\cellcolor[HTML]{C0C0C0} Figure & \cellcolor[HTML]{C0C0C0} Depth & \cellcolor[HTML]{C0C0C0} Crawling & \cellcolor[HTML]{C0C0C0}Standing & \cellcolor[HTML]{C0C0C0} Sitting  \\ \hline
\ref{fig:features_poses} i.&	1&	\cellcolor {yellow!50}0.33& 0&\cellcolor{green!25}	0.67  \\ \hline
	&2&	 \cellcolor{green!25}0.83&	0.001&\cellcolor {yellow!50}	0.169  \\ \hline
	&3&	0&	 \cellcolor{green!25}1&	0  \\ \hline
	&4&	\cellcolor{green!25}0.63&	0&\cellcolor {yellow!50}	0.37 \\ \hline
	&5&\cellcolor {yellow!50}0.20& 0&\cellcolor{green!25}	0.80  \\  \hline \hline
\ref{fig:features_poses} ii.&	1&	0& \cellcolor{green!25}	1&	0  \\ \hline
	&2&	\cellcolor {yellow!50}0.01& \cellcolor{green!25}	0.99&	0.001  \\ \hline
	&3&	0& \cellcolor{green!25}	1&	0  \\ \hline
	&4&	0& \cellcolor{green!25}	1&	0  \\ \hline
	&5&	0& \cellcolor{green!25}	1&	0  \\ \hline \hline
\ref{fig:features_poses} iii.	&1& \cellcolor{yellow!50}0.33&	0.02& \cellcolor{green!25}	0.65   \\ \hline
	&2&	 \cellcolor{green!25}0.536&	\cellcolor {yellow!50}0.311&	0.153  \\ \hline
	&3&	0& \cellcolor{green!25}	1&	0  \\ \hline
	&4&	 \cellcolor{green!25}0.62&	0.097&\cellcolor {yellow!50}	0.283   \\ \hline
	&5&	 \cellcolor{green!25}.8&	0&\cellcolor {yellow!50}	0.2   \\ \hline 
\end{tabular}
\end{table}

\begin{table}[]
\centering
\caption{Prediction score for Fire classification for different depths}
\label{tab:PredictionFire}

\begin{tabular}{|l|l|r|r|}
\hline
\cellcolor[HTML]{C0C0C0} Figure & \cellcolor[HTML]{C0C0C0} Depth & \cellcolor[HTML]{C0C0C0} Fire & \cellcolor[HTML]{C0C0C0} No Fire  \\ \hline
\ref{fig:features_fire} i.&	1&	0&	 \cellcolor{green!25}1  \\ \hline
	&2&	0.001&  \cellcolor{green!25}	0.999  \\ \hline
	&3&	 \cellcolor{green!25}1&	0  \\ \hline
	&4&	0&	 \cellcolor{green!25}1  \\ \hline
	&5&	0&	 \cellcolor{green!25}1  \\ \hline \hline
\ref{fig:features_fire} ii.&	1& \cellcolor{green!25}	1& 0  \\ \hline
	&2&	 \cellcolor{green!25}1&	0  \\ \hline
	&3&	 \cellcolor{green!25}1&	0  \\ \hline
	&4&	 \cellcolor{green!25}1&	0  \\ \hline
	&5&	 \cellcolor{green!25}0.999&	0.001  \\ \hline
\end{tabular}
\end{table}
\subsection{Accuracy and precision}

The accuracy of the neural network for different depths is consistent with the visualization of the features and the t-SNE. Figures \ref{fig:F1Scores} and \ref{fig:precision} show the F1 scores and the precision of the network at depths 1 to 5. For the purposes of this research, the networks have been trained to detect objects pertinent to fire navigation and rescue including doors, people, ladders, windows and combination of firefighters and windows.  The network with one layer(depth 1) has a reasonable accuracy close to 75\% and shows a high variance in the results. The depth 1 framework requires a low computational burden in training and test. Computational burden increases with the growth in computational complexity. Each additional depth increases computational complexity due to the additional number of Floating Points(FLOPs) operations. Interestingly, the use of 2 and 3 depths produces an unacceptable performance, but dramatically improves when using 4 and 5 depths. The use of  depth-5 is not necessary since its performance is almost identical to the run with 4 layers but the computational burden is significantly higher. 
\begin{figure}
\begin{center}
\includegraphics[scale=0.1]{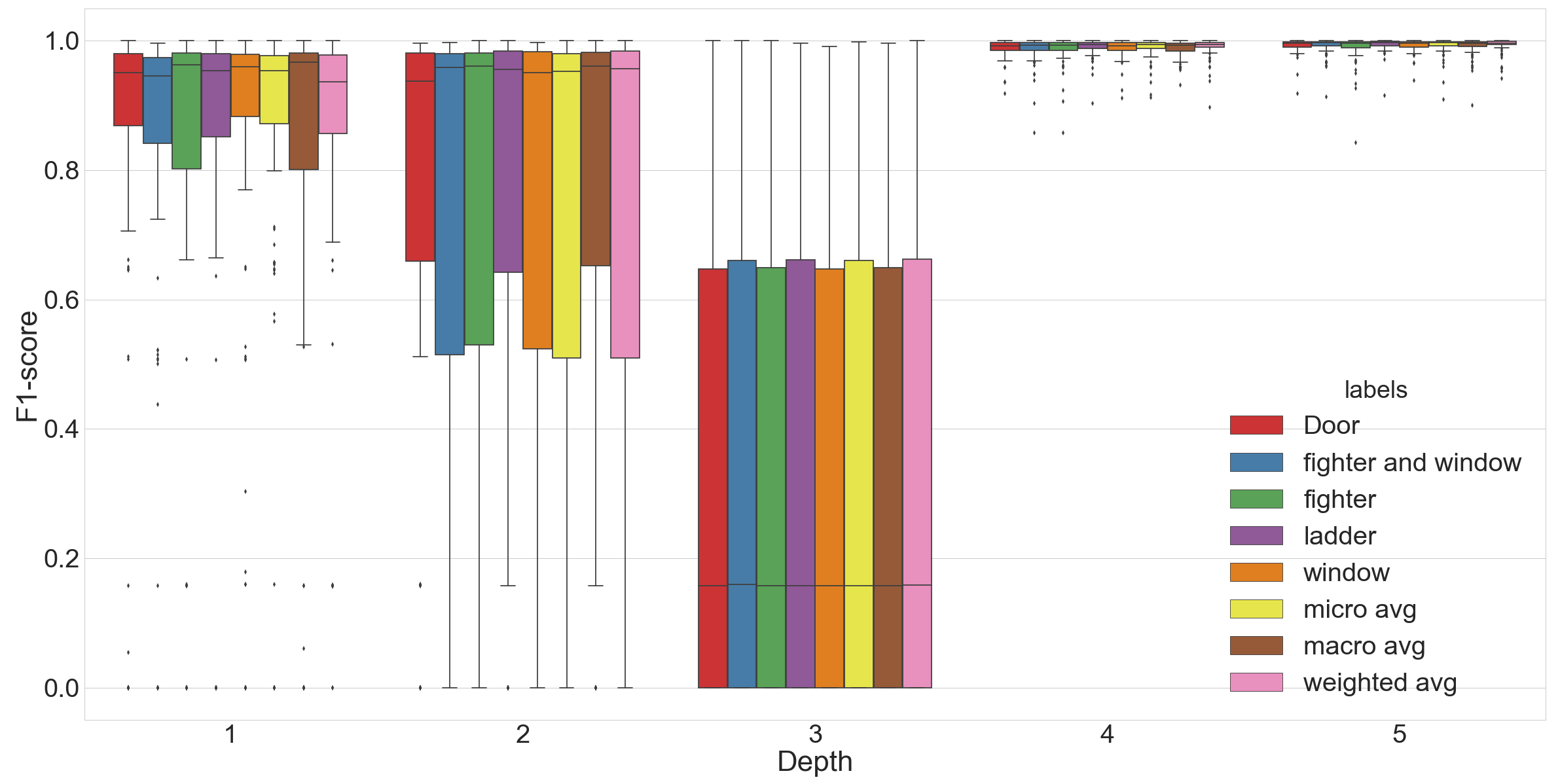}
\caption{F1 scores for the classification of objects with CNNs of different depths.}\label{fig:F1Scores}
\end{center}
\end{figure}

\begin{figure}
\begin{center}
\includegraphics[scale=0.1]{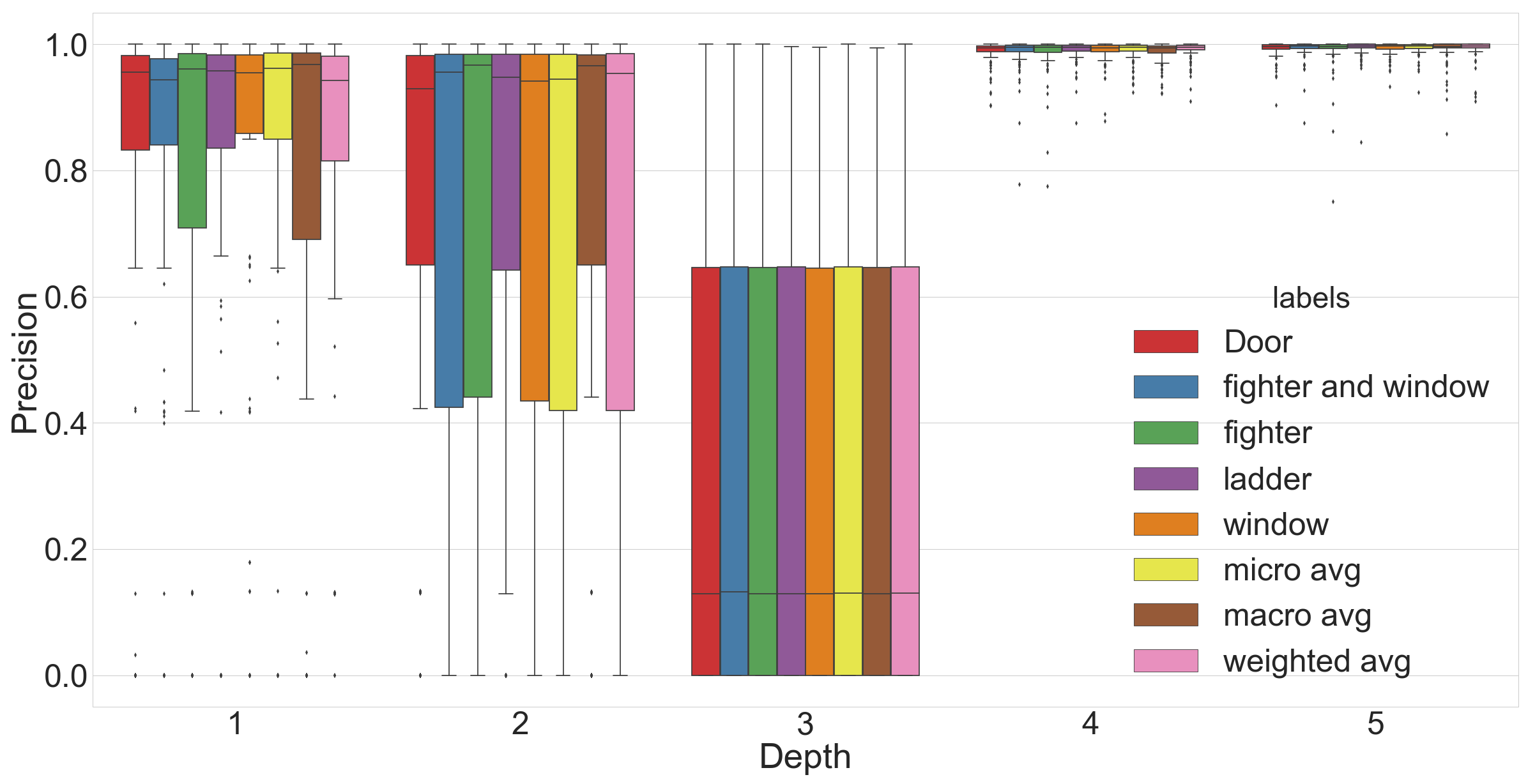}
\caption{Precision curves for object detection with CNNs of different depths.}\label{fig:precision}
\end{center}
\end{figure}

\begin{figure}
\begin{center}
\includegraphics[scale=0.1]{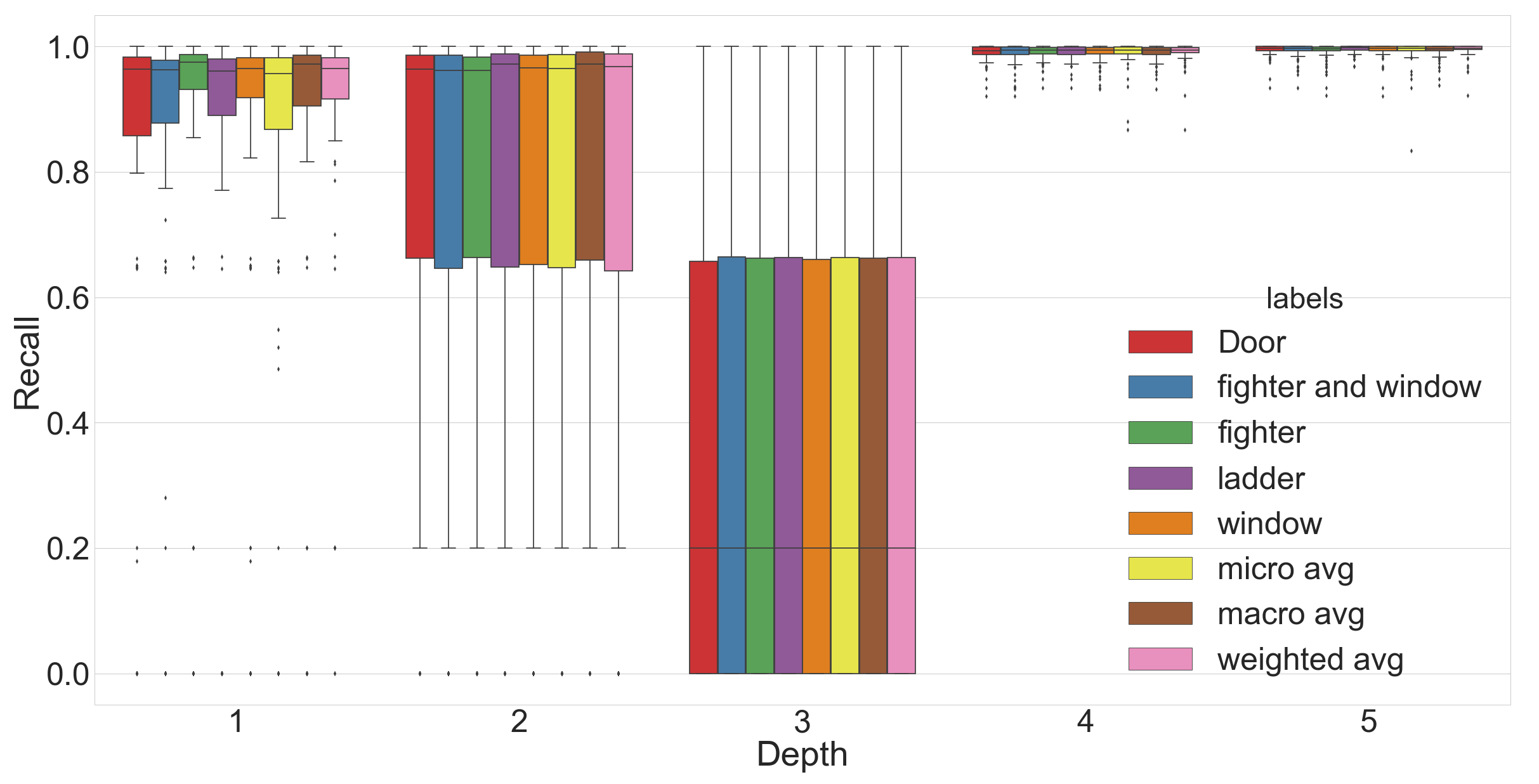}
\caption{Recall curves for object detection with CNNs of different depths.}\label{fig:precision}
\end{center}
\end{figure}

\begin{figure}
\begin{center}
\includegraphics[scale=0.1]{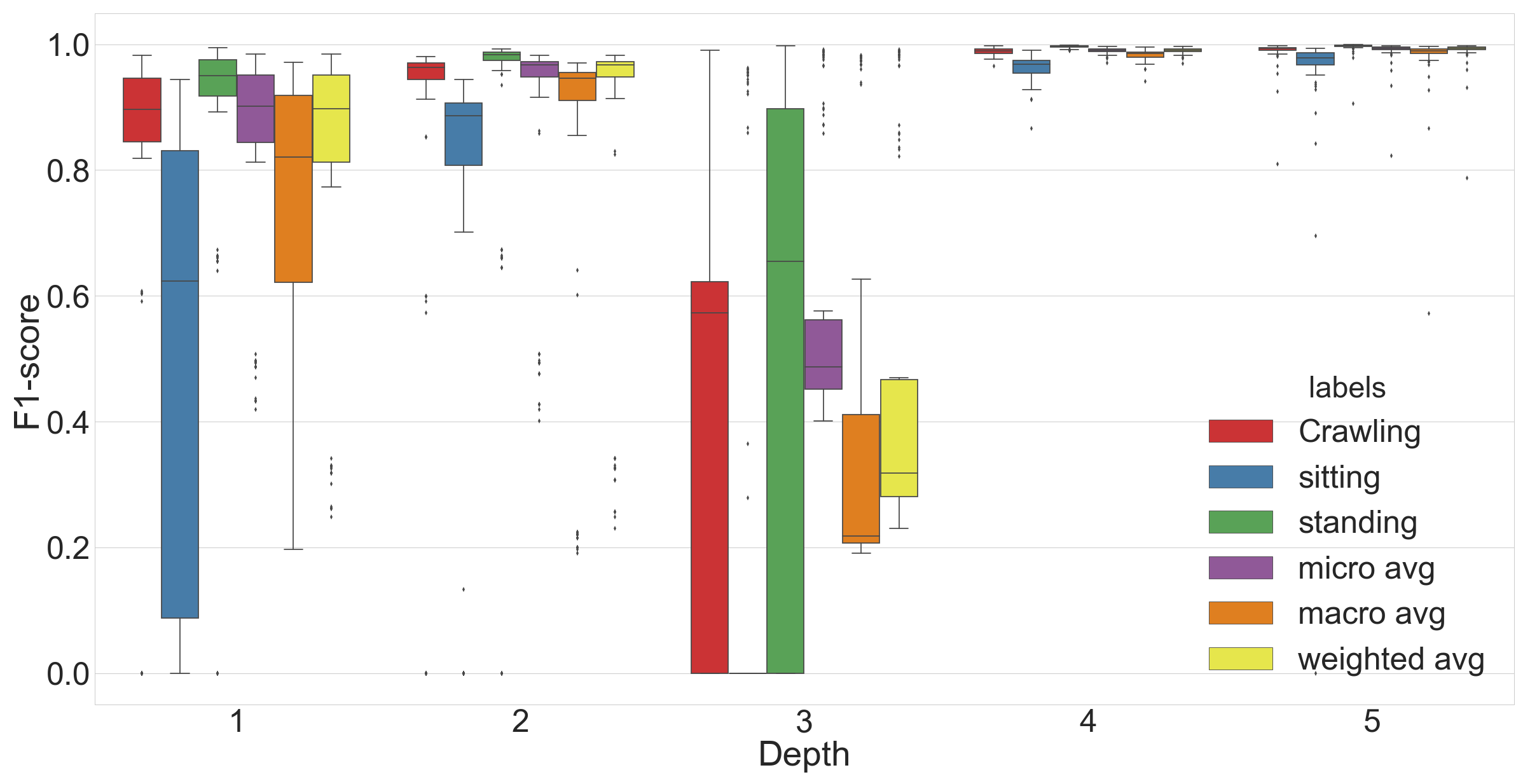}
\caption{F1 scores for the classification of poses with CNNs of different depths.}\label{fig:F1Scores}
\end{center}
\end{figure}

\begin{figure}
\begin{center}
\includegraphics[scale=0.1]{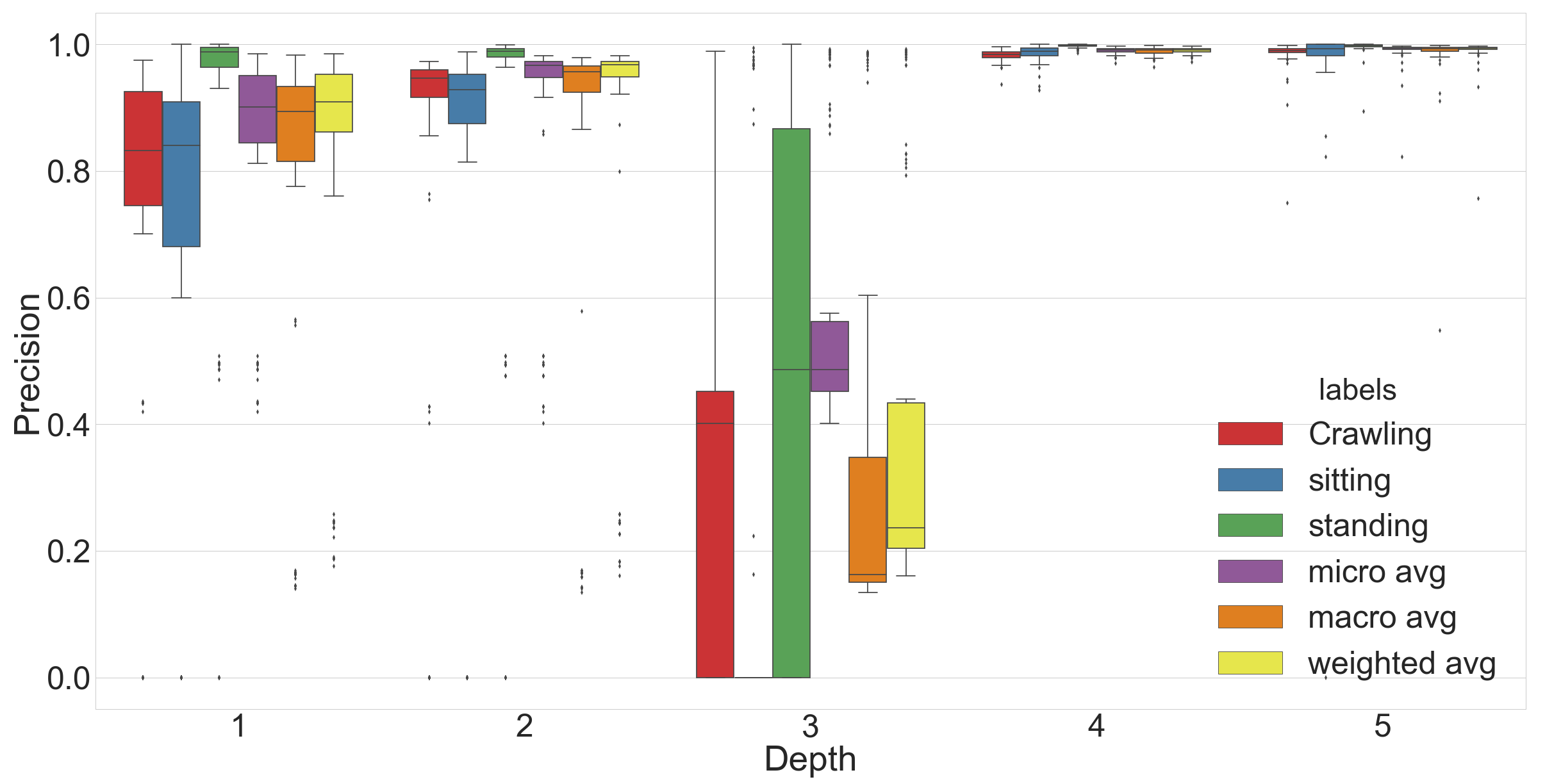}
\caption{Precision curves for poses detection with CNNs of different depths.}\label{fig:precision}
\end{center}
\end{figure}

\begin{figure}
\begin{center}
\includegraphics[scale=0.1]{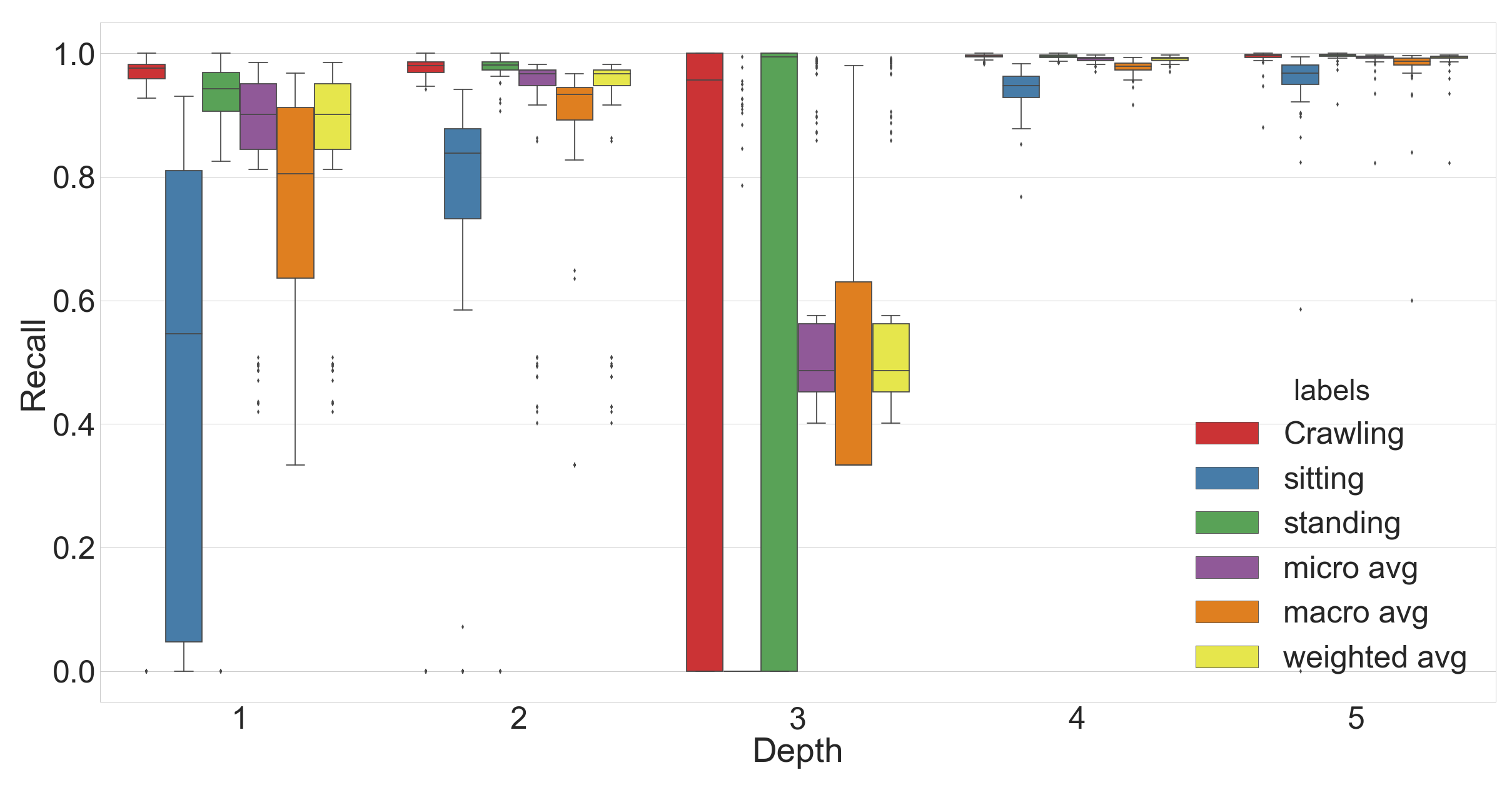}
\caption{Recall curves for poses detection with CNNs of different depths.}\label{fig:precision}
\end{center}
\end{figure}

\begin{figure}
\begin{center}
\includegraphics[scale=0.08]{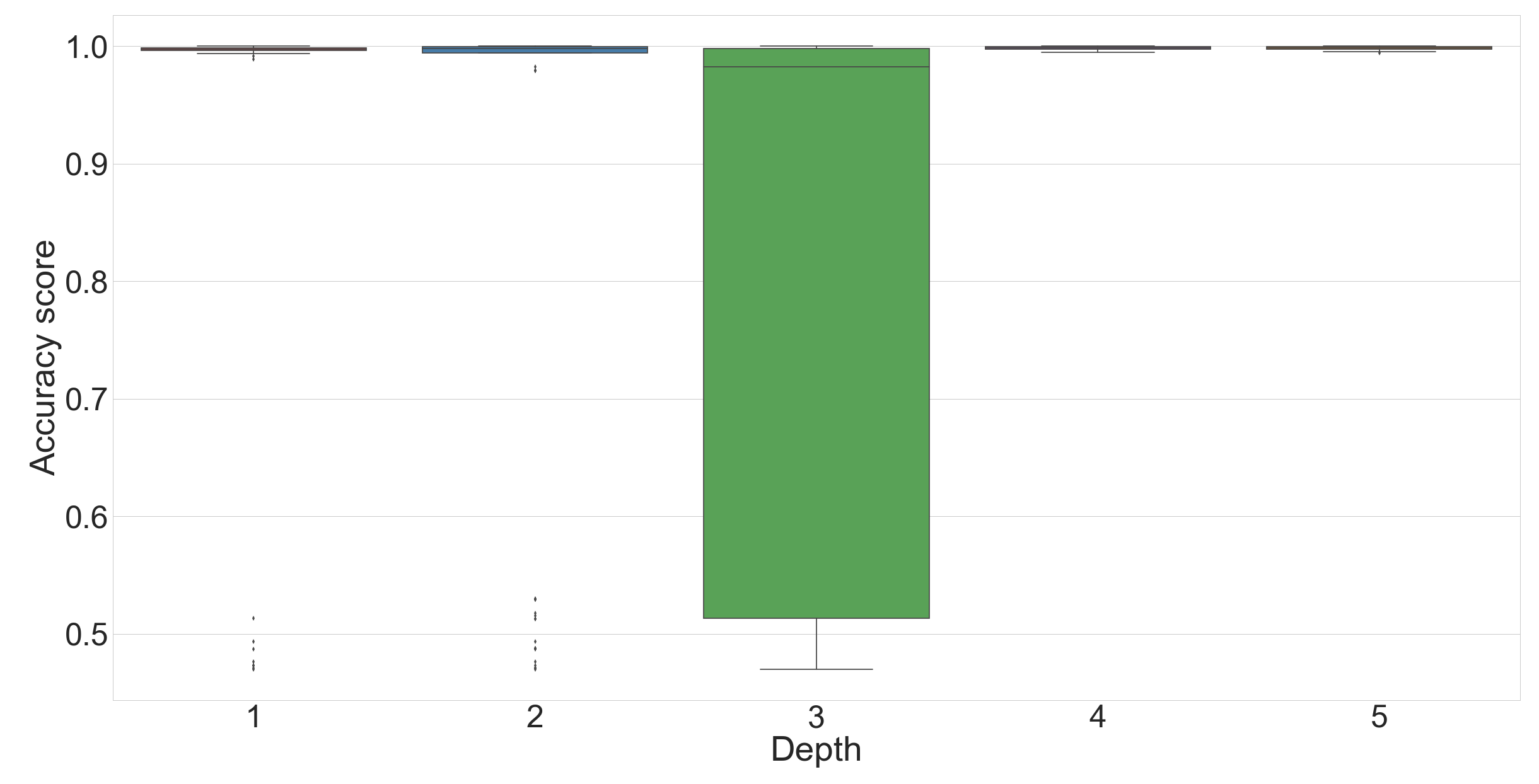}
\caption{Test accuracy in fire detection with CNNs of different depths.}\label{fig:fireAccuracy}
\end{center}
\end{figure}

\begin{figure}
\begin{center}
\includegraphics[scale=0.08]{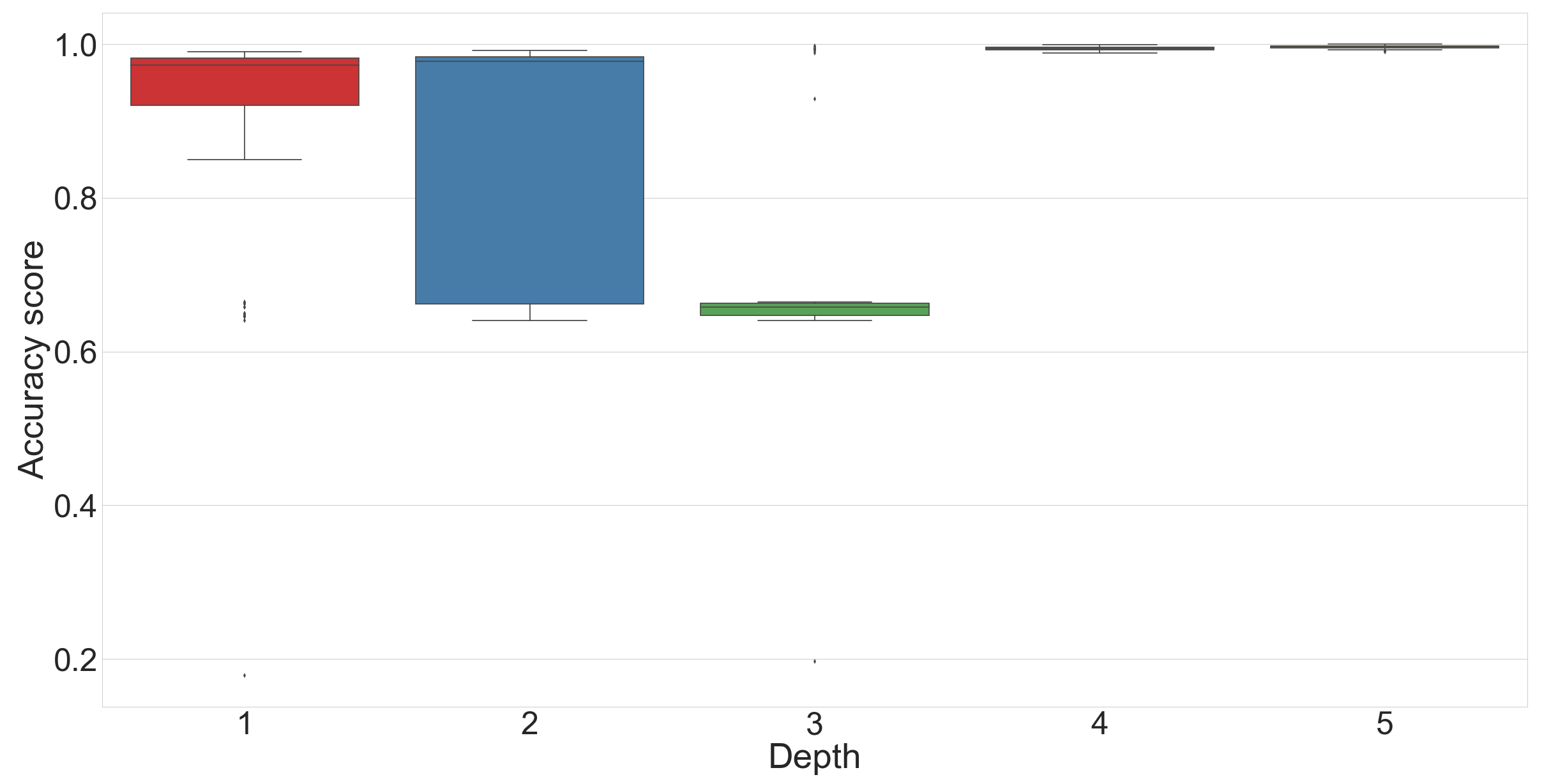}
\caption{Test accuracy in object detection with CNNs of different depths.}\label{fig:testAccuracy}
\end{center}
\end{figure}

Figure \ref{fig:fireAccuracy} shows the accuracy of the network when detecting fire. In this case it is only necessary to run the network with one layer. As in the previous experiments, the network with three layers produces a poor performance, while the other depth tests show an accuracy that is almost identical. The average test accuracy for all objects is depicted in figure \ref{fig:testAccuracy}. From this graph, we can conclude that a good trade off between computational burden and accuracy is obtained with a network of only one convolutional layer, while the highest level of accuracy is obtained using a network with 4 layers, which can achieve an accuracy of more than 97\%. 

The network's ability to accurately distinguish between differing human positions and classify them accordingly presents new and interesting opportunities. With pose recognition, body position can be used to assist in making health inferences. For example, the presence of a person laying down is very important in a fire scenario, as it may be a significant indicator of a person who has succumbed to smoke inhalation and is in desperate need of rescue. The results can aid in alerting rescue teams to the possible health condition of victims and prioritize evacuation. In our tests, we trained the network to detect persons standing, sitting or crawling from labelled images. The network shows a similar performance with accuracy of 95\% on average. 

\begin{figure}
\begin{center}
\includegraphics[scale=0.09]{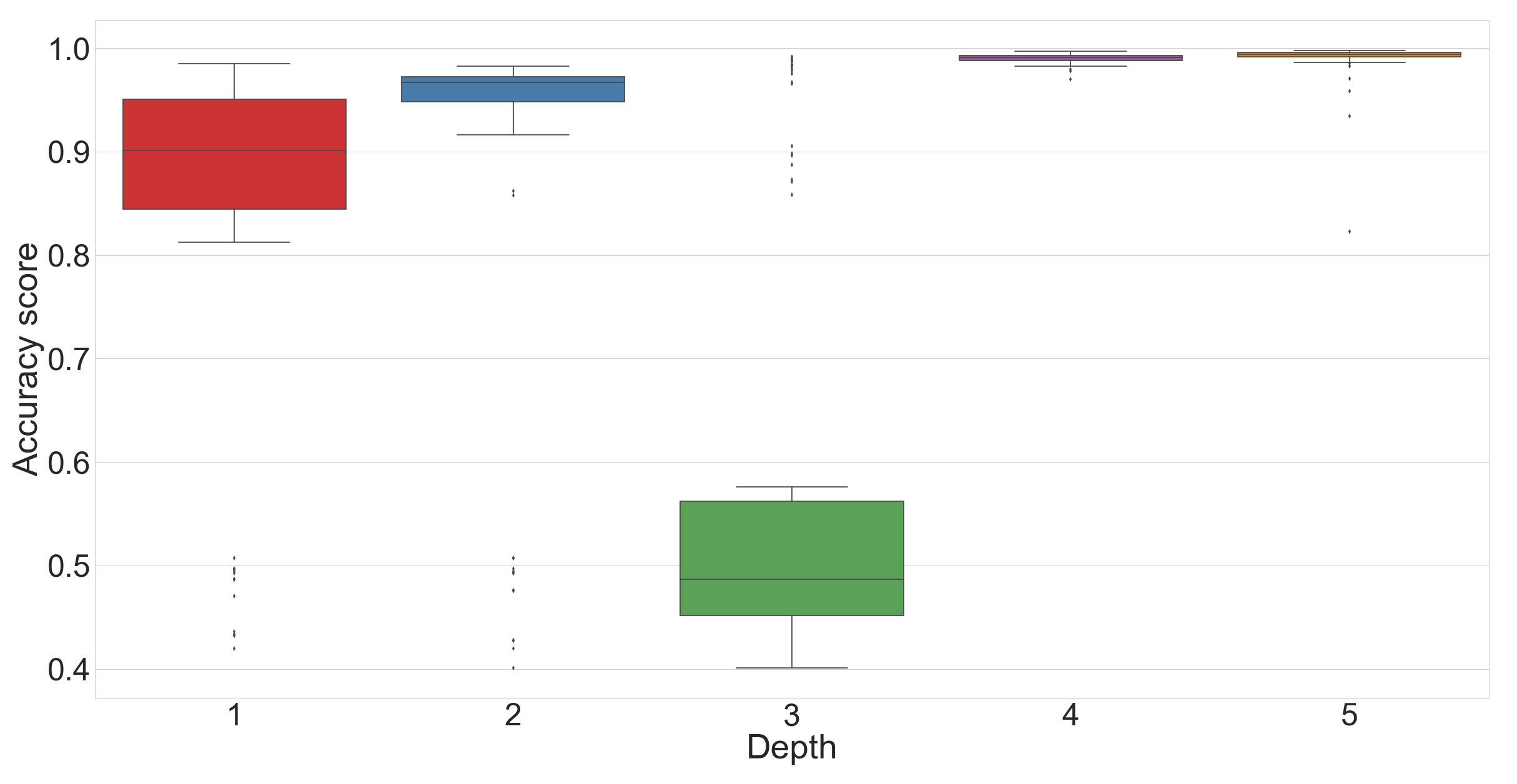}
\caption{Test accuracy in pose detection with CNNs of different depths.}
\end{center}
\end{figure}

\subsection{Confusion matrices}
The confusion matrices for all detection modalities have been computed. Tables \ref{tab:confusionObjects1} to \ref{tab:confusionObjects5} shows the confusion matrices for the task of object and human detection. As stated before, the network with just one depth of convolutional blocks performs reasonably well without excess computational burden. The detection probabilities are around 75\%. We note however that at lower depths(i.e 1 and 2) the network is challenged to make clear distinctions between classes that tend to occur together. For example, both windows and ladders tend to occur in a scene with one or more firefighters. The network has a high confusion rate in determining windows and ladders as individual classes, as there is a confusion of about 19\% with firefighters in the classification of these elements. This is likely due to a significantly higher number of unaugmented images in the firefighter class when compared to the others in the "object of interest" set. When the number of depths is increased to 2, the confusion is even worse, reaching 35\%. However, accuracy of detection of firefighters as a class is between 97 and 99\%. If the number of layers is increased to 4, the confusion matrices show a confusion between 0 and 1.7\% for all objects being classified. As stated above, there are no significant differences between 4 and 5 depths. Note that the classification performance with these number of depths ranges between 97 and 99.8\%.

\begin{table}[]
\centering
\caption{Confusion matrix for object detection, depth=1}
\label{tab:confusionObjects1}
\begin{tabular}{|l|r|r|r|r|r|}
\hline
Door & \cellcolor[HTML]{C0C0C0}73.2 & 1.1 & 23.5 & 0.0 & 2.2 \\ \hline
F/W & 0.0 & \cellcolor[HTML]{C0C0C0}77.7 & 19.0 & 0.0 & 3.3 \\ \hline
Fighter & 0.3 & 1.8 & \cellcolor[HTML]{C0C0C0}97.0 & 0.1 & 0.8 \\ \hline
Ladder & 0.0 & 1.2 & 19.1 & \cellcolor[HTML]{C0C0C0}79.6 & 0.1 \\ \hline
Window & 0.3 & 4.3 & 19.9 & 0.0 & \cellcolor[HTML]{C0C0C0}75.5 \\ \hline
 & \multicolumn{1}{l|}{Door} & \multicolumn{1}{l|}{F/W} & \multicolumn{1}{l|}{Fighter} & \multicolumn{1}{l|}{Ladder} & \multicolumn{1}{l|}{Window} \\ \hline
\end{tabular}
\end{table}

\begin{table}[]
\centering
\caption{Confusion matrix for object detection, depth=2}
\label{tab:confusionObjects2}
\begin{tabular}{|l|l|l|l|l|l|}
\hline
Door & \cellcolor[HTML]{C0C0C0}61.3 & 0.0 & 38.2 & 0.0 & 0.5 \\ \hline
F/W & 0.0 & \cellcolor[HTML]{C0C0C0}62.4 & 36.0 & 0.0 & 1.7 \\ \hline
Fighter & 0.1 & 0.2 & \cellcolor[HTML]{C0C0C0}99.4 & 0.1 & 0.2 \\ \hline
Ladder & 0.0 & 0.0 & 34.3 & \cellcolor[HTML]{C0C0C0}65.7 & 0 \\ \hline
Window & 0.1 & 1.0 & 35.9 & 0 & \cellcolor[HTML]{C0C0C0}63.0 \\ \hline
 & Door & F/W & Fighter & Ladder & Window \\ \hline
\end{tabular}
\end{table}

\begin{table}[]
\centering
\caption{Confusion matrix for object detection, depth=4}
\label{tab:confusionObjects4}
\begin{tabular}{|l|l|l|l|l|l|}
\hline
Door & \cellcolor[HTML]{C0C0C0}97.1 & 0.2 & 0.9 & 0.0 & 1.7 \\ \hline
F/W & 0.0 & \cellcolor[HTML]{C0C0C0}98.3 & 1.0 & 0.0 & 0.7 \\ \hline
Fighter & 0.1 & 0.1 & \cellcolor[HTML]{C0C0C0}99.8 & 0.0 & 0.0 \\ \hline
Ladder & 0.0 & 0.0 & 0.2 & \cellcolor[HTML]{C0C0C0}99.8 & 0.0 \\ \hline
Window & 0.3 & 0.5 & 0.3 & 0.0 & \cellcolor[HTML]{C0C0C0}98.9 \\ \hline
 & Door & F/W & Fighter & Ladder & Window \\ \hline
\end{tabular}
\end{table}

\begin{table}[]
\centering
\caption{Confusion matrix for object detection, depth=5}
\label{tab:confusionObjects5}
\begin{tabular}{|l|l|l|l|l|l|}
\hline
Door & \cellcolor[HTML]{C0C0C0}98.0 & 0.0 & 0.7 & 0.0 & 1.3 \\ \hline
F/W & 0.0 & \cellcolor[HTML]{C0C0C0}99.2 & 0.5 & 0.0 & 0.3 \\ \hline
Fighter & 0.0 & 0.1 & \cellcolor[HTML]{C0C0C0}99.9 & 0.0 & 0.0 \\ \hline
Ladder & 0.0 & 0.0 & 0.1 & \cellcolor[HTML]{C0C0C0}99.9 & 0.0 \\ \hline
Window & 0.3 & 0.6 & 0.2 & 0.0 & \cellcolor[HTML]{C0C0C0}98.9 \\ \hline
 & Door & F/W & Fighter & Ladder & Window \\ \hline
\end{tabular}
\end{table}

A similar trend in detection accuracy related to the number of depths can be seen in the detection of poses. Results at depths 1 and 2 perform poorly compared to 4 and 5 as shown in Tables \ref{tab:confusionPoses1},  \ref{tab:confusionPoses2}, \ref{tab:confusionPoses4} and  \ref{tab:confusionPoses5} . At depths 1 and 2, the network was mainly challenged in distinguishing the differences between sitting and crawling. This is possibly due to the fact that the relative positions of the firefighters in these two poses are similar but simply rotated. The accuracy increases significantly with 4 depths (Table \ref{tab:confusionPoses4}), where the confusion decreases to a range between 0 and 0.4 in all cases except for the detection of the sitting pose, which stands at a confusion rate of 5.7 (in the network with 4 depths) and 5.1\% with the crawling pose with 5 depths (Table \ref{tab:confusionPoses5}). Again we found no significant differences between the results of these two network configurations.   

\begin{table}[]
\centering
\caption{Confusion matrix for pose detection, depth=1}
\label{tab:confusionPoses1}
\begin{tabular}{|l|r|r|r|}
\hline
Crawling & \cellcolor[HTML]{C0C0C0}85.6 & 1.2 & 13.2 \\ \hline
Sitting & 41.4 & \cellcolor[HTML]{C0C0C0}45.9 & 12.7 \\ \hline
Standing & 10.8 & 0.4 & \cellcolor[HTML]{C0C0C0}88.8 \\ \hline
 & \multicolumn{1}{l|}{Crawling} & \multicolumn{1}{l|}{Sitting} & \multicolumn{1}{l|}{Standing} \\ \hline
\end{tabular}
\end{table}

\begin{table}[]
\centering
\caption{Confusion matrix for pose detection, depth=2}
\label{tab:confusionPoses2}
\begin{tabular}{|l|l|l|l|}
\hline
Crawling & \cellcolor[HTML]{C0C0C0}86.3 & 0.8 & 12.9 \\ \hline
Sitting & 19.3 & \cellcolor[HTML]{C0C0C0}67.8 & 12.9 \\ \hline
Standing & 7.5 & 0.2 & \cellcolor[HTML]{C0C0C0}92.3 \\ \hline
 & Crawling & Sitting & Standing \\ \hline
\end{tabular}
\end{table}

\begin{table}[]
\centering
\caption{Confusion matrix for pose detection, depth=4}
\label{tab:confusionPoses4}
\begin{tabular}{|l|l|l|l|}
\hline
Crawling & \cellcolor[HTML]{C0C0C0}99.5 & 0.2 & 0.3 \\ \hline
Sitting & 5.7 & \cellcolor[HTML]{C0C0C0}94.1 & 0.2 \\ \hline
Standing & 0.5 & 0.0 & \cellcolor[HTML]{C0C0C0}99.5 \\ \hline
 & Crawling & Sitting & Standing \\ \hline
\end{tabular}
\end{table}

\begin{table}[]
\centering
\caption{Confusion matrix for pose detection, depth=5}
\label{tab:confusionPoses5}
\begin{tabular}{|l|l|l|l|}
\hline
Crawling & \cellcolor[HTML]{C0C0C0}99.4 & 0.2 & 0.4 \\ \hline
Sitting & 5.1 & \cellcolor[HTML]{C0C0C0}94.7 & 0.2 \\ \hline
Standing & 0.5 & 0.0 & \cellcolor[HTML]{C0C0C0}99.5 \\ \hline
 & Crawling & Sitting & Standing \\ \hline
\end{tabular}
\end{table}

\subsection{Detection  versus false alarm probabilities}

\begin{figure}
\begin{center}
\includegraphics[scale=0.12]{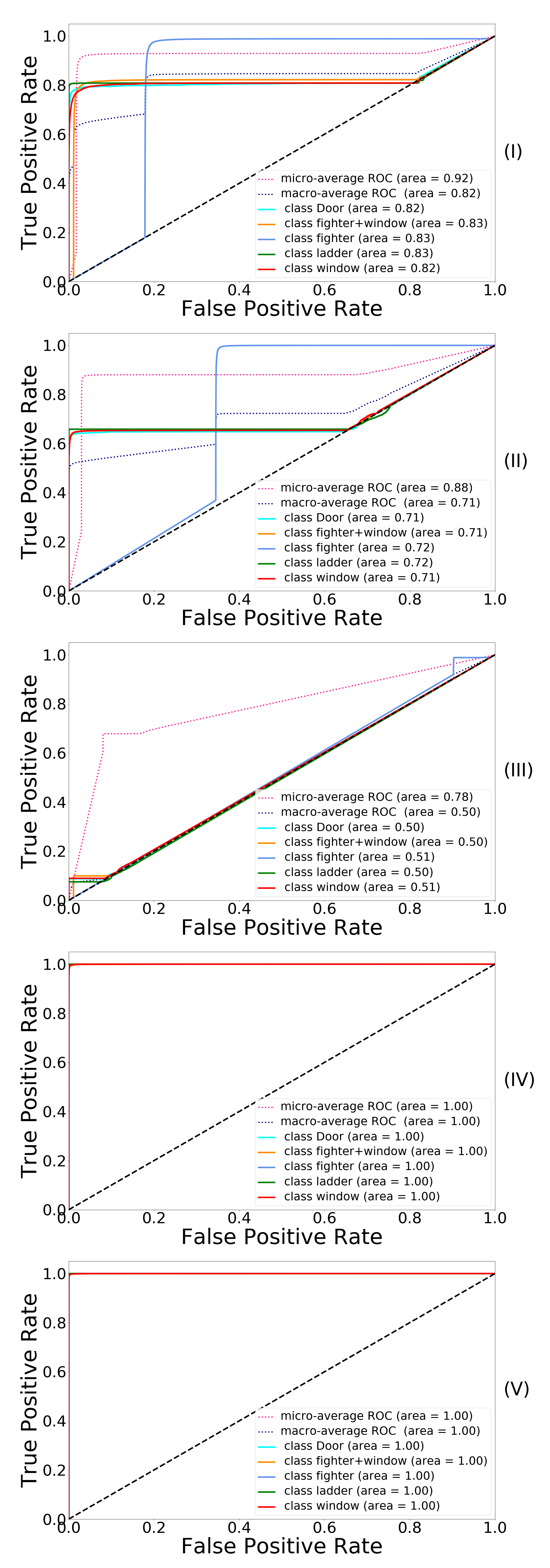}
\caption{ROC curve for object detection for I) Depth-1, II) Depth-2, III) Depth-3, IV) Depth-4 and V) Depth-5 architectures. }\label{fig:ROC_object}
\end{center}
\end{figure}

\begin{figure}
\begin{center}
\includegraphics[scale=0.12]{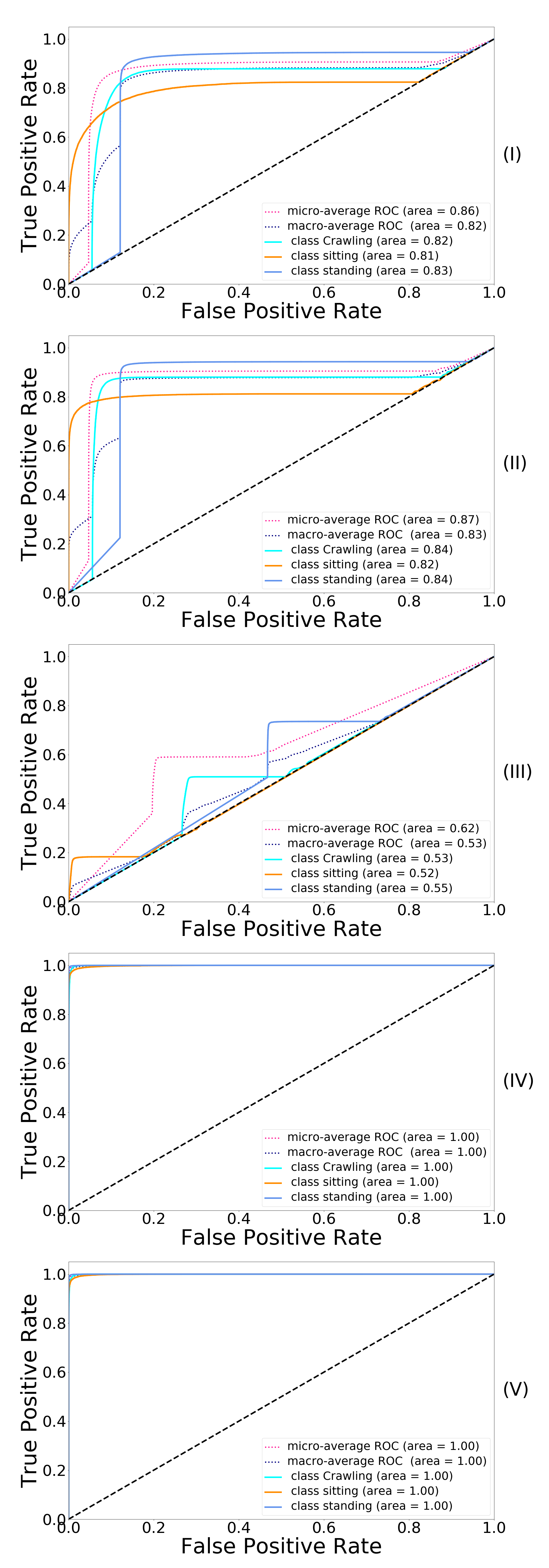}
\caption{ROC curve for pose detection for I) Depth-1, II) Depth-2, III) Depth-3, IV) Depth-4 and V) Depth-5 architectures .}\label{fig:ROC_pose}
\end{center}
\end{figure}

\begin{figure}
\begin{center}
\includegraphics[scale=0.12]{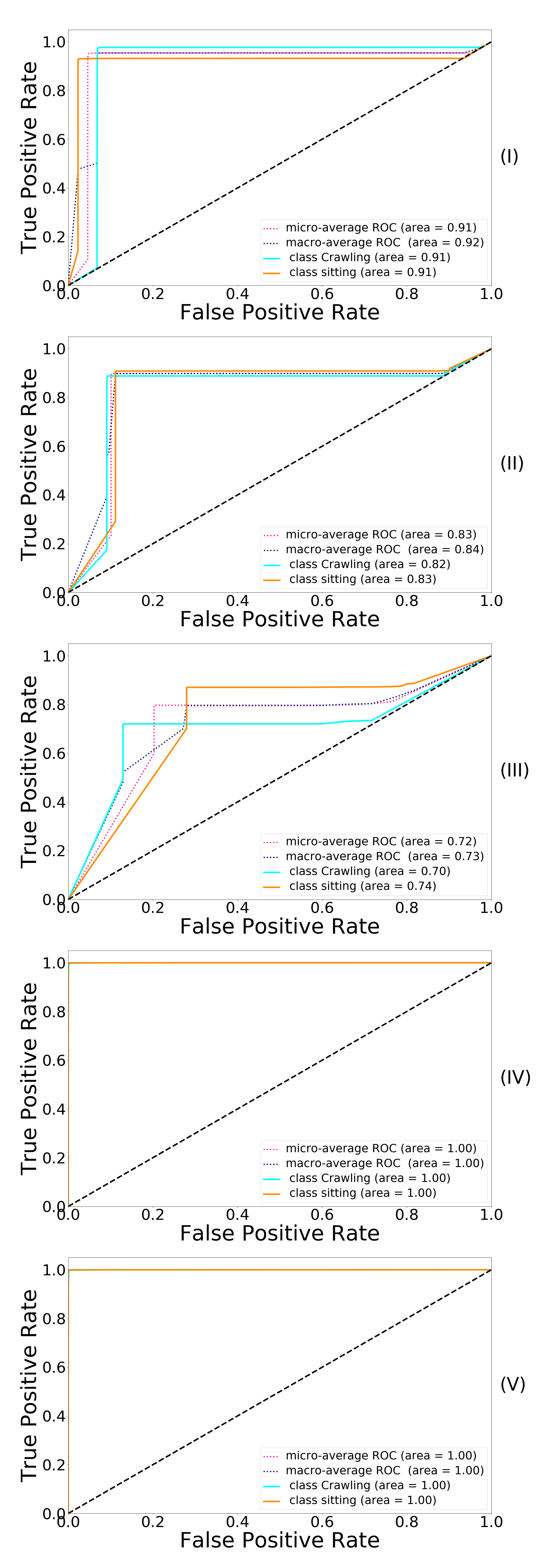}
\caption{ROC curve for fire detection for I) Depth-1, II) Depth-2, III) Depth-3, IV) Depth-4 and V) Depth-5 architectures .}\label{fig:ROC_fire}
\end{center}
\end{figure}

Figures \ref{fig:ROC_object}, \ref{fig:ROC_pose} and \ref{fig:ROC_fire}  show the probability of detection versus false alarm for the classification of objects and humans, poses and fire. The graphs have been obtained by sweeping the detection threshold from 0 to 1. In all cases, it is observed that the probability of false alarm is negligible when the networks have a depth of 4 or 5. As expected given the poor performance of the classifier at depths 2 and 3, the results showed high false alarm probabilities. In the case of the depth 1 network, the results show flaws in the probability of detection versus probability of false alarm. A detection rate of about 82\% is achieved with a false alarm rate of about 18\% (\ref{fig:ROC_object})for object detection as per the micro/macro average ROC area. The optimal false alarm rates in individual class object detection range from 15-20\%, in particular, in the detection of humans. Similar results are obtained in pose detection. Therefore, although depth 1 networks can be used with lower computational burden, they should only be used in cases where such false alarm rates can be tolerated. False alarm probability can be decreased if the detection is performed on a sequence of images chronologically, and then a voting procedure is applied to all detections. This would be done at the cost of additional computational time in the object or pose detection procedures.

\section{Conclusion}

The scene of a structural fire is chaotic, dangerous, and disorienting. Heavy smoke, near zero visibility, extreme heat and active flames create a perfect storm for stress induced misjudgments that can affect even seasoned fire fighters. These are prime conditions for computer aided assistance and artificially intelligent solutions. Our research provides a mechanism that can supplement firefighters with real time information and offer guidance by automatically interpreting the fireground from the information provided by the hand held thermal cameras already in use by firefighters.

We present a deep learning based technology that is capable of accurate automated detection of objects of interest utilizing thermal imagery being recorded on the scene. Convolutional neural networks have demonstrated outstanding performance in object detection on RGB imagery. However, the zero light, heavy smoke conditions typical of structural fires render RGB cameras useless. The CNN developed for this application is unique in that it achieves high detection accuracy applied to thermal imagery and is capable of processing, analysis and result generation in real time if a camera is connected to a simple single board commercial computer endowed with GPU capabilities (for example, an NVIDIA JETSON). Our model is able to accurately detect objects of interest such as doors, windows, and ladders vital to evacuation. It is also able to detect people and differentiate postures. Posture detection may assist in prioritization of rescue as a rough estimate of the health status of a victim (for instance a prone posture may indicate a person who has succumbed to the effects of smoke inhalation and require immediate evacuation and paramedic assistance). Our framework is capable of performing with greater than 95\% accuracy on the detection of these objects of interest and more then 90\% accuracy on posture identification. We also present an evaluation of computational time to accuracy achievement trade-off and show that the model performs above 70\% even at the lowest convolutional depth, making it highly adept to usage in the field where computational power may be a concern. 

This work lays a foundation to the development of a real time situational map of the structural configuration of a building that is actively built and updated via the live thermal imagery being recorded by firefighters moving through the scene. This map, which is updated in real time, could be used by firefighters to assist them in safely navigating the burning structure and improve the situational awareness necessary in decision making by tracking exits that may become blocked and finding alternatives. Utilizing the features detected via the approach presented in this paper, a robust localization and tracking system to track objects of interest in sequences of frames can be built. The visual features from this framework have also be coupled with a Natural Language Processing(NLP) system for scene description and allow the framework to autonomously make human understandable descriptions of the environment to aid firefighters to improve their understanding of the immediate surroundings and assist them when anxiety levels are heightened. Our future work seeks to join these two components  with a reinforcement learning(Q-learning) algorithm that utilizes the continuously updated state map and reinforcement learning techniques that assist in path planning and can be vocalized through the NLP system. The deep Q-learning  based approach provides a navigation system that actively avoids hazardous paths. These three components, built on the backbone of the research presented here, can be fused to accomplish the ultimate goal of providing an artificially intelligent solution capable of guiding firefighters to safety in worst case scenarios.


\bibliographystyle{cas-model2-names}

\bibliography{cas-refs}


\newpage
\bio{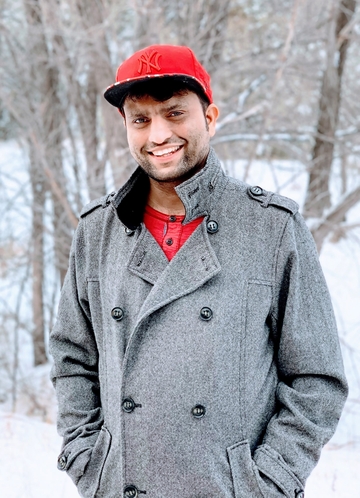}
Manish Bhattarai received the Masters degree in Electrical Engineering with specialization in signal processing from the University of New Mexico in 2017.He is currently working towards his Ph.D. degree in Electrical Engineering from the same school as well as employed as a full time research assistant at the Los Alamos National Laboratory.  His research interests are  Machine Learning, deep learning ,Computer Vision, AI and HPC. 
\endbio

\bio{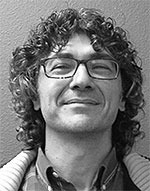}

Manel~Mart\'{i}nez-Ram\'on is a professor with the ECE department of The University of New Mexico. He holds the King Felipe VI Endowed Chair of the University of New Mexico, a chair sponsored by the Household of the King of Spain. He is a Telecommunications Engineer (Universitat Politecnica de Catalunya, Spain, 1996) and PhD in Communications Technologies (Universidad Carlos III de Madrid, Spain, 1999). His research interests are in Machine Learning applications to smart antennas, neuroimage, first responders and other cyber-human systems, smart grid and others. His last work is the monographic book "Signal Processing with Kernel Methods",  Wiley, 2018.
\endbio

\end{document}